# Machine learning for early prediction of circulatory failure in the intensive care unit

Stephanie L. Hyland[1,2,3,4,+], Martin Faltys[5,+], Matthias Hüser[1,4,+], Xinrui Lyu[1,4,+], Thomas Gumbsch[6,8,+], Cristóbal Esteban[1,4], Christian Bock[6,8], Max Horn[6,8], Michael Moor[6,8], Bastian Rieck[6,8], Marc Zimmermann[1], Dean Bodenham[6,8], Karsten Borgwardt[6,8, ‡], Gunnar Rätsch[1,2,3,4,7,8,‡], Tobias M. Merz[5,9,‡]

1 Department of Computer Science, ETH Zürich, Zürich, Switzerland
2 Computational Biology Program, Memorial Sloan Kettering Cancer Center, New York, U.S.A.
3 Tri-Institutional PhD Program in Computational Biology and Medicine, Weill Cornell Medicine, New York, U.S.A.
4 Medical Informatics Group, Zürich University Hospital, Zürich, Switzerland
5 Department of Intensive Care Medicine, University Hospital, and University of Bern, Bern, Switzerland
6 Department of Biosystems and Engineering, ETH Zürich, Basel, Switzerland
7 Department of Biology, ETH Zürich, Zürich, Switzerland
8 Swiss Institute for Bioinformatics, Lausanne, Switzerland
9 Cardiovascular Intensive Care Unit, Auckland City Hospital, Auckland, New Zealand

+ Joint-first authors
‡ These authors jointly directed this work. Contact: tobiasm@adhb.govt.nz, gunnar.ratsch@ratschlab.org, karsten.borgwardt@bsse.ethz.ch

## Abstract

Intensive care clinicians are presented with large quantities of patient information and measurements from a multitude of monitoring systems. The limited ability of humans to process such complex information hinders physicians to readily recognize and act on early signs of patient deterioration. We used machine learning to develop an early warning system for circulatory failure based on a high-resolution ICU database with 240 patient-years of data. This automatic system predicts 90.0% of circulatory failure events (prevalence 3.1%), with 81.8% identified more than two hours in advance, resulting in an area under the receiver operating characteristic curve of 94.0% and area under the precision recall curve of 63.0%. The model was externally validated in a large independent patient cohort.

Critical illness is characterized by the presence or risk of developing life-threatening organ dysfunction. Critically ill patients are typically cared for in intensive care units (ICU), which are designed to provide the highest level of patient observation and support. ICUs specialize in providing continuous monitoring and advanced therapeutic and diagnostic technologies such that organ function can be maintained while the underlying illness or injury is treated. An integral aspect of patient care is regular patient evaluation, including assessing changes in organ function parameters over time and in the context of established treatments. Many ICUs now use electronic patient data management systems (PDMS) to centrally store and manage patient-specific data, and enable easily-interpretable display of trend data on bedside monitors. Collected data includes measurements of organ function parameters with high temporal resolution, results of diagnostic tests, and parameters determined by therapeutic interventions.

ICU physicians are presented with large and growing quantities of data from many patients and it is increasingly difficult to identify the most important information for care decisions. The limited ability of humans to process such quantities of information can lead to data overload, change blindness, and task fixation[1]. Critical care settings are therefore vulnerable to problems related to failures by clinicians to readily recognize, interpret, and act upon relevant information[2,3], leading to delays in care provision. Crucially, in a majority of patients the causes of deterioration are potentially treatable and most interventions will be more efficient if initiated early[4–6]. A common approach to identify patients at risk of organ system deterioration is the use of medical device alarms for individual physiological measurements based on critical thresholds. Since such alarm systems lack access to comprehensive information their alarms are often nonspecific[7,8] and lead to alarm fatigue, which was rated seventh on the list of top 10 technology hazards from the ECRI Institute[9,10]. Machine learning techniques excel in the analysis of complex signals in data-rich environments[11,12]. Recent advances in deep learning, reinforcement learning, and other machine learning techniques have enabled and popularized their use in mathematics[13], engineering[14], biology[15] and medicine[16–26,77]. The abundance of data collected in the ICU and the importance of timely decision-making are key to the growing interest of using of machine learning in this setting.

The potential of data analysis using modern machine learning methods is not well established in a ICU context[29]. With some exceptions[28,30,31], only simple statistical models for detection or prediction of specific conditions have been developed and deployed[32–38]. Despite being based on relatively few features and including far fewer information categories than those available to clinicians, these models showed acceptable prediction results for the examined prediction problems. In the machine learning community, likely spurred by the availability of datasets such as MIMIC-III[39], intensive care has attracted significant research interest. Endpoints such as patient mortality[40] and length of stay (LOS)[41] are commonly tackled using predictive models. However, after the initial decision to admit a patient to ICU for full treatment the accurate prediction of mortality or LOS is not of great importance for further treatment decisions. Patient state deterioration besides mortality has also been addressed in the context of ICU admission[42] or the onset of treatment[43]. Prediction of circulatory failure has not been extensively explored, with existing work focusing primarily on specific aspects such as hypotension[44] and vasopressor use[45], or related problems like sepsis prediction[28,46,47] and renal failure[30].

In this work, we develop a novel approach based on medical knowledge, large-scale data analysis, and state-of-the-art machine learning techniques to construct two early warning systems for circulatory failure in ICU patients. These systems, named *circEWS* (/ˈsərkəs/) and *circEWS-lite*, are of differing complexity and alert clinicians to patients at risk of circulatory failure in the next 8 hours. We define patients as being in circulatory failure if they a) have elevated lactate (≥ 2 mmol/L), and b) have either low blood pressure (MAP≤ 65 mmHg) or are receiving vasopressors or inotropes. To train the early warning system, we use a large database from a multidisciplinary ICU in a tertiary care center in Berne, Switzerland, containing physiological variables, diagnostic test results, and treatment parameters routinely collected from more than 54,000 ICU admissions. With close to 5,000 variables collected and continuously monitored parameters updated every 2 minutes, this dataset contains over 3 billion data points. It has higher temporal resolution compared to the two publicly available ICU datasets (MIMIC-III[39] contains around 312 million data points and eICU[48] contains around 827 million) and non-ICU datasets[19,20,22], allowing better characterization of patient states[49]. We have developed a comprehensive analysis framework including data pre-processing and cleaning, feature extraction and interpretation, and a selection of large-scale supervised machine learning techniques to construct *circEWS*.

To evaluate the performance of *circEWS*, we established an alarm/event-based evaluation measure, which assesses the fraction of circulatory failure events correctly predicted (i.e., an alarm was raised for this event) and the false alarm rate (i.e., there was an alarm but no event). For external validation we applied different versions of our system to the MIMIC-III database.

## Results

### *Preparation of a high time resolution ICU data set (HiRID)*

The full dataset contained a total of 4,959 different routinely collected physiological variables, diagnostic test results, and treatment parameters from 54,225 patient admissions to the ICU (Supplementary Fig. 1 shows an example). After applying exclusion criteria (Supplementary Fig. 2a,b), information from 710 variables concerning 36,098 patient admissions during the period from January 2008 to June 2016 remained for further processing (Supplementary Tab. 1 shows a summary of patient characteristics). After merging redundant variables and inclusion of static patient characteristics, 209 consistently measured variables were used for model development (Fig. 1a, Supplementary Fig. 2c). This was achieved by aggregating pharmaceutical variables across administration routes and dosages, and summarizing measurement modalities of physiological variables (Supplementary Fig. 4a). The data was resampled to 5-minute resolution using adaptive imputation for missing measurements (Fig. 1b). The patient's circulatory state was annotated for each time point as "circulatory failure", "no circulatory failure", or "ambiguous" (insufficient data for annotation) (Fig. 1c). Overall, we identified 45,886 instances of circulatory failure in 11,046 patients with a mean event duration of 320 minutes. We found that ICU mortality correlated with longer duration and higher number of events of circulatory failure (Supplementary Fig. 3). We aim to release the pre-processed data to the research community on Physionet[78] in order to enhance reproducibility and enable academic re-use of this dataset.

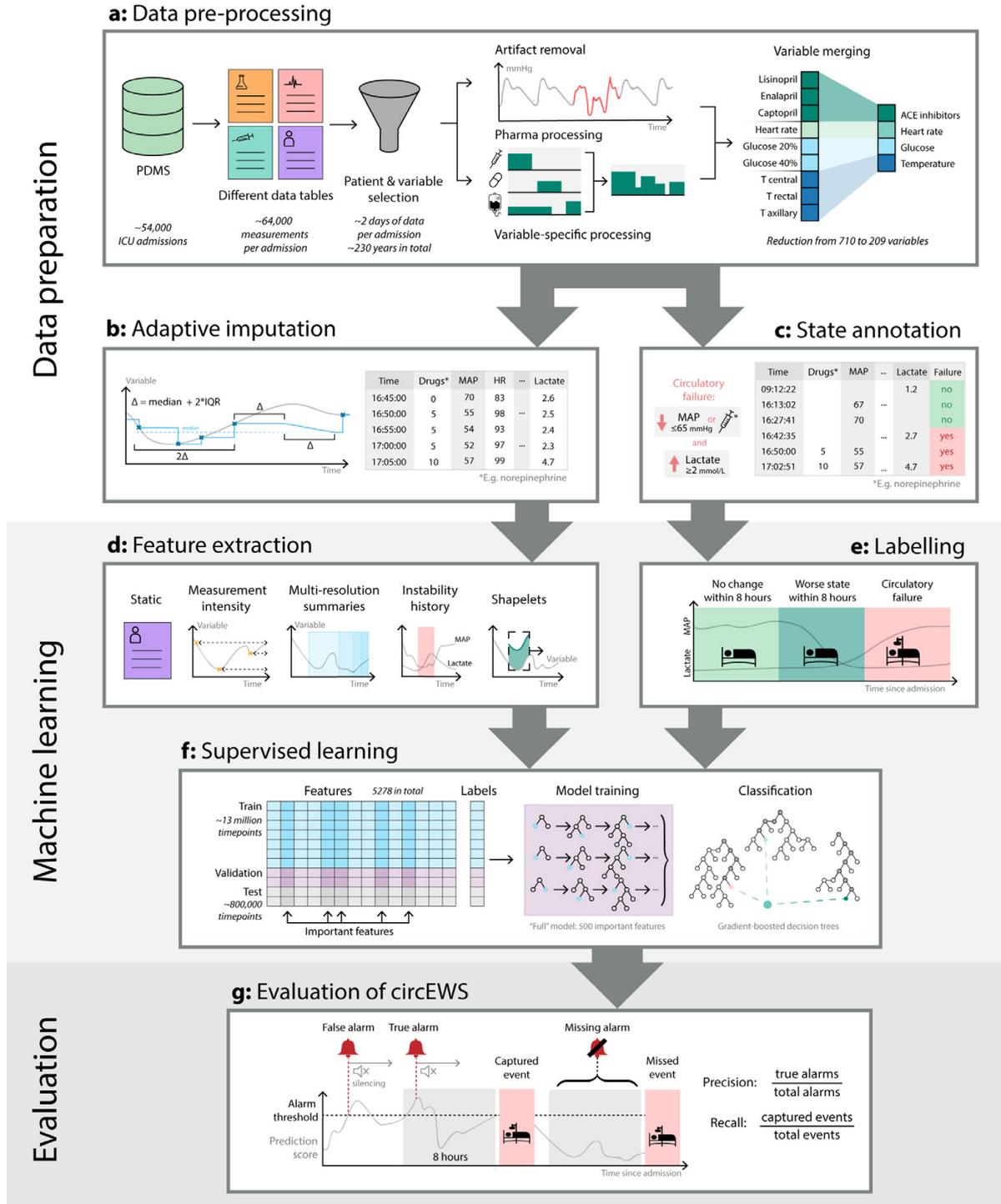

**Fig. 1: Model development overview.**

**Data preparation: a:** Data on patient admissions were exported from the ICU patient database management system (PDMS) and filtered according to inclusion/exclusion criteria. Relevant variables for the prediction problem were identified. Clinically implausible values, variable-specific errors, and other artefacts were automatically excluded using variable-specific algorithms. Pharmaceutical variables determined by differing administration methods were merged to produce effective administration rates over time. Different monitoring modalities for the same parameter were merged. **b:** Adaptive imputation was performed by resampling the data to a 5-minute time grid and filling missing values using a patient and variable-specific imputation scheme. **c:** The circulatory state was annotated according to the circulatory failure definition. **Machine learning: d:** At each time-point, four feature types (intensity, multi-resolution summaries, instability history, shapelets) per variable were extracted and together with static patient information represented the data available at this time point. **e:** To construct the binary prediction problem, each relevant time-point was labelled as either "positive" (circulatory failure occurs in the next 8 hours) or "negative" (circulatory failure does not occur in the next 8 hours). **f:** A binary classifier to predict near-term circulatory failure was trained on the extracted labels and features. A gradient-boosted ensemble of decision trees was chosen as the classifier after benchmarking different methods. **Evaluation: g:** The proposed early warning system for circulatory deterioration, called *circEWS,* consists of the trained binary classifier, a decision threshold, and a policy of silencing for a short period after alarms. The system was evaluated based on the fraction of alarms which are correct (precision), and the fraction of circulatory failure events which are correctly predicted (recall).

### *Development of a machine-learning framework for prediction of circulatory failure*

Using the ***HiRID*** data, we constructed a two-class dataset for machine learning based on the patient's near-future circulatory state. All time points annotated as "no circulatory failure" (i.e., not currently in a circulatory event state) were labelled as "positive" if circulatory failure occurs in the next 8 hours, otherwise "negative" (Fig. 1e). Ambiguously labelled time points were excluded from training and evaluation. We extracted a high-dimensional feature vector per time point, which included static patient information, multi-scale summaries of time series history, measurement intensity of variables, and temporal patterns (Fig. 1d). Overall, the dataset contained 15 million labelled time slices of which 3.1% were labelled positive. With all available imputed features/labels the dataset was ≈300 GB in size. We developed an analysis framework that is capable of dealing with such large datasets and supports feature selection, model selection, and evaluation of temporal datasets. Based on an advanced experimental setup and evaluation strategy (Supplementary Fig. 4b), we tested and compared multiple state-of-the-art machine learning approaches, including deep learning. Among different classifiers, gradient-boosted ensembles of decision trees (*lightGBM*[50]) were found to offer the best overall performance (Supplementary Fig. 5d,e).

Two classifiers with differing complexity were developed - "full" and "compact" models. From the 710 variables in the dataset, 5,278 features were constructed and ranked according to mean absolute *SHAP* value[51] which indicates their importance for predictions on the validation set. The "full" model uses the top 500 features, originating from 112 variables (Supplementary Tab. 5). We then determined commonly available variables that were ranked within the top 20 according to mean absolute *SHAP* value (Table 1; specifically, we require the variable to be available in MIMIC-III). This lead to a list of 17 variables that produce 176 features which are used in the "compact" model. As a baseline, a decision tree model was developed, which uses only the last measurement of each of the three variables included in the definition of circulatory failure (MAP, lactate level, and dosage of vasopressors/inotropes). This baseline model mimics a traditional threshold-based rule system based on a small number of variables. The areas under the receiver operator curves (AUROCs) with respect to the time point-based labeling of the full, compact, and baseline models were 94.0%, 93.9%, and 88.3%, respectively (Fig. 2a). The areas under the precision recall curves (AUPRCs) are more informative for rare events and were 46.7%, 45.4%, and 25.4% for full, compact, and baseline models, respectively (Supplementary Fig. 5a).

| Rank (std) | Variable | Important feature categories |
|---|---|---|
| 1 (0.0) | Lactate | Current, Shapelet, Multi-resolution, Instability history, Measurement |
| 2 (0) | MAP | Multi-resolution, Instability history, Current, Shapelet, Measurement |
| 3 (5.3) | Time since ICU admission | N/A |
| 4 (0.4) | Patient age | Static |
| 5 (3.0) | Heart rate | Current, Multi-resolution, Measurement, Shapelet |
| 6-9 (2.3) | Dobutamine, Milrinone, Levosimendan*, Theophylline* | Instability history, Multi-resolution |
| 10 (5.3) | Cardiac output | Shapelet, Multi-resolution, Measurement |
| 11 (3.5) | RASS | Current, Multi-resolution, Measurement |
| 12 (34.6) | INR | Measurement, Multi-resolution, Current |
| 13 (5.8) | Serum glucose | Multi-resolution, Current, Measurement |
| 14 (4.4) | C-reactive protein | Multi-resolution, Current, Measurement |
| 15 (7.9) | Diastolic BP | Multi-resolution, Shapelet, Measurement |
| 16 (4.0) | Peak inspiratory pressure (Ventilator) | Current, Measurement, Multi-resolution, Shapelet |
| 17 (7.9) | Systolic BP | Current, Multi-resolution, Measurement, Shapelet |
| 18 (10.6) | SpO$_2$ | Multi-resolution, Shapelet, Measurement |
| 19 (17.8) | Non-opioid analgesics* | Multi-resolution |
| 20 (11.4) | Supplemental oxygen | Multi-resolution, Measurement, Current |

**Table 1:** Table of the top 20 ranked variables for the prediction of circulatory failure. The ranking was obtained by first ranking all 5,278 features according to their importance in explaining predictions of the development model and then greedily selecting clinical variables in a forward-selection procedure if they contribute to important features derived from these variables. The point estimate of the rank is obtained on the held-out data split, and the standard deviation of the rank was obtained on five splits of the data. The last column lists the important feature categories for a variable, i.e., the feature categories that contribute to the top 50 features overall. The categories are sorted by decreasing importance in terms of rank in the list of top features. MAP: Mean arterial pressure, BP: Blood pressure, RASS: Richmond Agitation Sedation Scale, INR: International Normalized Ratio (Prothrombin time). Variables not contained in MIMIC III are marked by * and were not used in the compact model (and hence also not the *circEWS-lite* system) as they appear to be less commonly available.

***The circulatory early warning system (circEWS)***

Our model generates a prediction score every 5 minutes that is associated with the risk of the patient experiencing circulatory failure in the next 8 hours. A simple, threshold-based warning system derived from this prediction score could lead to a continuous stream of alarms every 5 minutes, causing alarm fatigue. We therefore developed a more advanced alarm system that triggers an alarm when the likelihood for circulatory failure exceeds a specified threshold. It also implements a silencing policy: once an alarm is triggered, subsequent alarms are then suppressed for 30 minutes (Fig. 1g). If the patient experiences circulatory failure and recovers during the silencing period, the system is "reset" and unsilenced to potentially inform clinicians about recurrence of circulatory failure. The effects of different silencing periods and reset times are shown in Supplementary Fig. 6. We applied this alarm algorithm to the predictions of our full and compact model, and called the resulting systems *circEWS* and *circEWS-lite,* respectively. The performance of the two systems is shown in Fig. 2b, using precision-recall curves[52] (PRC) to illustrate different trade-offs of precision and recall. Recall was defined as the fraction of events with *any* alarm in the preceding eight hours (excluding the five minutes immediately prior to the event) and precision as the fraction of alarms which correctly predicted an event. Precision and recall measure performance on the raised alarms and occurring events, respectively, and are clinically more meaningful than time point-based measures. It should be noted, that for rare events (only 3.1% of time points are labelled positive), predictions with high precision are much more difficult to obtain than small false positive rates. Recall as a function of time before occurrence of circulatory failure for fixed overall recall and precision is shown in Fig. 2e. We observe an increase in the rate of correctly predicted events closer to the onset of circulatory failure, with 81.8% of the events identified more than 2 hours in advance. The timeliness of *circEWS alarms* is further illustrated in Fig. 2f, showing the temporal distribution of the first alarm and number of alarms in the 8-hour window prior to deterioration. The results imply that, if a medical practitioner with an 8-hour shift was to react to every alarm from *circEWS*, every patient would be checked only every second shift and on average 2 hours 32 minutes before circulatory failure (Fig. 2f).

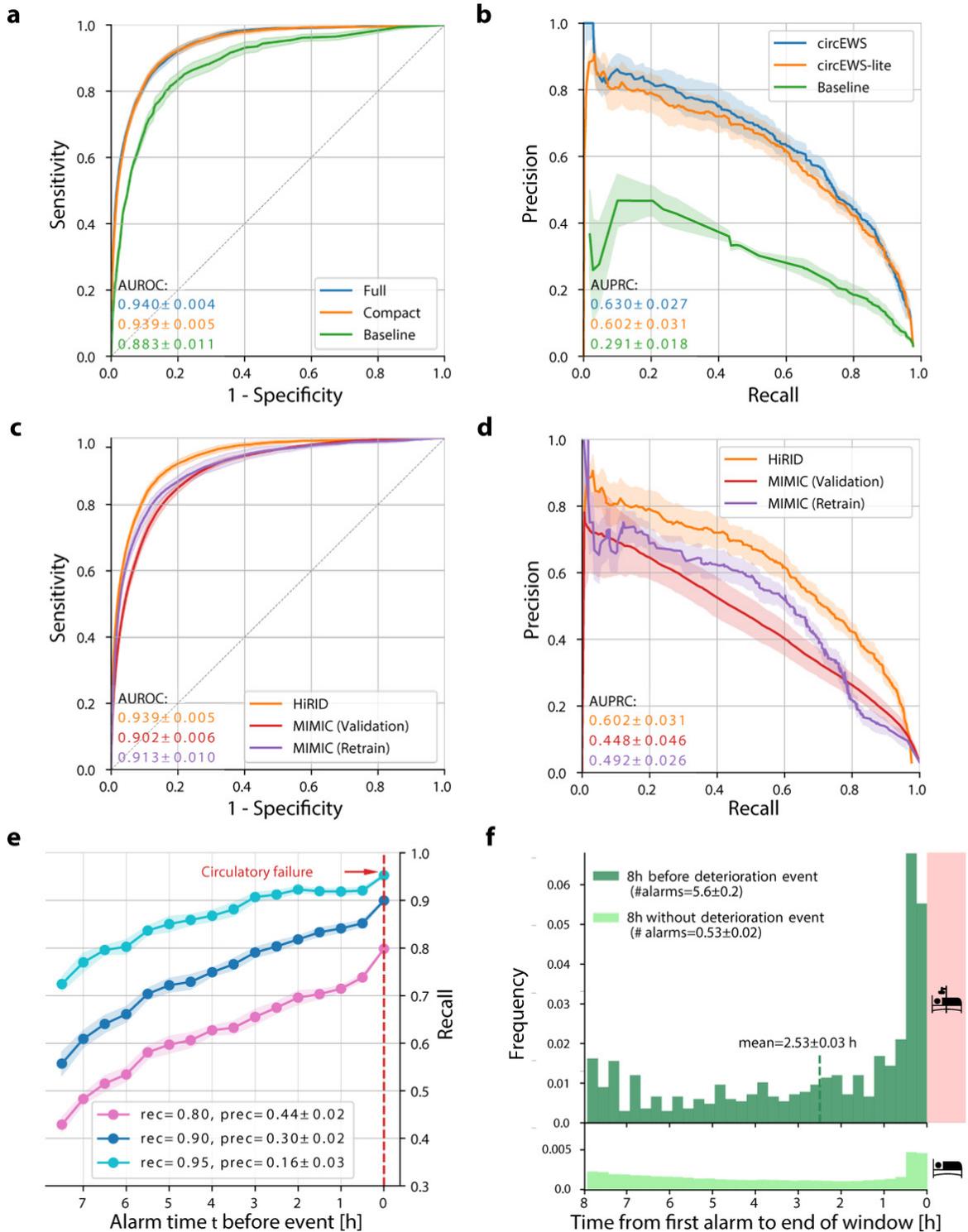

**Fig. 2: Model performance. a:** Receiver-operating characteristic (ROC) curve for the binary classification task of predicting circulatory failure, comparing our two proposed models to a baseline model. The full model contained 500 features (composed from 112 variables) and the compact model contained 176 features (composed from 17 variables). The baseline model only used variables included in the definition of circulatory failure and is based on a decision tree. **b**: Precision-recall curve for the *circEWS/circEWS-lite* alarm systems built on top of the full/compact classification model from panel a, using 30min silencing. Recall was defined as the fraction of events for which the system correctly raised an alarm from 8 hours to 5 min before the event. Precision was defined as the fraction of alarms which are in a window of 8 hours prior to a circulatory failure event. **c&d**: External validation: The compact model (*circEWS-lite*) was tested on the MIMIC-III ICU dataset and evaluated using ROC and PR curves, in analogy to panels a and b. *MIMIC (Validation)* denotes the performance

of the model trained on the HiRID dataset, and applied on the MIMIC-III data set, *MIMIC (Retrain)* denotes a model re-trained and tested on MIMIC-III data splits. Reported precision in MIMIC was corrected to reflect the different prevalence in MIMIC compared to HiRID **e**: Using a sliding 30-minute window, the fraction of events that are correctly retrieved by alarms as early as a certain time in (t,8] h before circulatory failure is reported. **f:** Shown is the distribution of the first alarm in the 8 hours before an alarm (top). The time to deterioration from the first alarm was on average 2 hours and 32 minutes. The bottom shows the distribution of raised alarms in 8 hour windows not immediatly followed by an event. We found the distribution is not uniform and hypothesize that this is due to past and future deterioration events (i.e., outside of the 8h period). In all panels, metrics are reported ± denoting standard deviation.

### *External validation*

The publicly available ICU dataset MIMIC-III[39] was used for external validation. The 17 variables required for *circEWS-lite* were identified in MIMIC (Table 1, Supplementary Table 5). We performed identical pre-processing of the MIMIC-III data with minor modifications to account for a lower time-resolution in MIMIC-III and different encodings. We report the performance of *circEWS-lite* on this MIMIC test set as *MIMIC (Validation)* in Fig. 2c,d. Additionally, model training was repeated on MIMIC; results are reported as *MIMIC (Retrain)*. In both cases, we corrected the label prevalence to be equal to the observed prevalence in HiRID, enabling comparison of precision in Fig. 2d (before correction MIMIC has 1.8% in contrast to HiRID's 3.1% positive labels). AUPRC results for uncorrected prevalence are shown in Supplementary Fig. 7c.

### *Inspection of model features*

We used *SHAP* values[51] to assess the importance of individual features of *circEWS*. For a given prediction, the influence of each feature on the classifier output is expressed as a *SHAP* value, with positive and negative *SHAP* values indicating an increase or decrease respectively, on the prediction score.

In Fig. 3a, we list the top 15 features by mean absolute *SHAP* value, and show the distribution of *SHAP* values across all predictions. Unsurprisingly, features from variables used to define circulatory failure rank highest. The relationship between feature value and *SHAP* value is illustrated in more detail for the features *Patient age* and *MAP* in Fig. 3b,c, with further examples in Supplementary Fig. 8. Table 1 reports the 20 most relevant variables (also used to define *circEWS-lite*). We analyzed the effect on the AUPRC of removing each of these variables in turn, and found that only the removal of lactate noticeably decreased performance (resulting AUPRC 41.1 ± 3.7%). Greedy forward selection of variables guided by performance on the validation set confirms lactate and MAP as the most important variables, as in the analysis based on *SHAP* values. Model performance begins saturating after adding around 10 variables (Supplementary Tab. 4). Fig. 3d shows the highest-ranking lactate shapelet as an example from the shapelet feature class, illustrating that the *SHAP* value of the feature increases 5.5 hours before onset of deterioration.

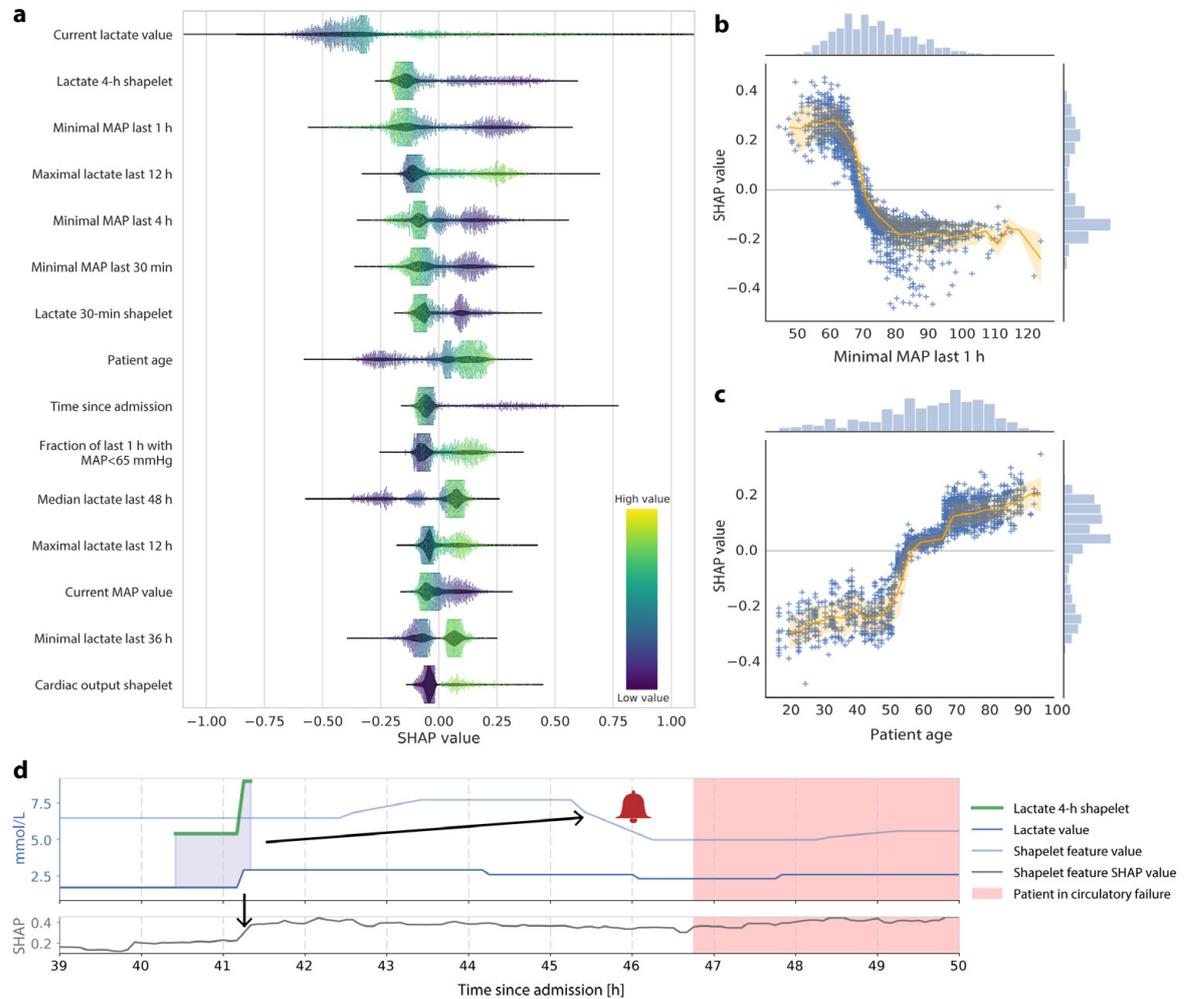

**Fig. 3: Feature inspection. a**: Top 15 features according to their mean absolute SHAP values. On the y-axis, the black violin plot shows the full distribution of the *SHAP* values for this feature. The dot plot in the foreground shows a colour-coding of the actual value of the feature resulting in the *SHAP* value on the x-axis. The color-coding is based on the percentile of the feature value in respect to the whole distribution. **b&c**: Scatter plots showing the relationship between feature value and *SHAP* value for age and lowest MAP in the last hour. The orange line represents the mean; standard deviation is given in shaded orange. The distributions of the *SHAP* values and feature values are shown as histograms on the right side and the top of the main graph. The high variance in the *SHAP* value for a given feature value indicates a strong influence of other features. **d**: Shapelet feature illustration: The lactate shapelet has an increase in its *SHAP* value (gray) 5.5 hours before the patient suffers from deterioration (first arrow). Shown in light blue is the feature value that represents the distance between time series and shapelet at 4 hours in the past. The feature value drops right before the event, increasing the prediction score (second arrow).

***Model performance in different patient cohorts***

In all subsequent analyses with fixed thresholds, we chose a threshold leading to a recall of 90.0%, resulting in a precision of 29.6% for circEWS. Results for the system with recalls of 80.0% and 95.0% are shown in Supplementary Fig. 10.

To study if the model performs similarly across different types of patients, we evaluated *circEWS* in different cohorts: varying age, severity of illness at the time of ICU admission, APACHE diagnostic groups, and compared medical and surgical patients as well as elective and emergency admissions (Fig. 4a-c, Supplementary Fig. 9c,e). We find similar performance across most diagnostic groups, with the exception of neurological patients for whom the model performs worse, with an event recall of 76.6% compared to 91.2% across all patients in the rest of the cohort (corrected p-value 0.038, dependent 2-sample t-test). For neurosurgical patients, the model exhibits a decreased precision of 8.1% compared with 30.0% in the rest of the patients (p-value 0.0006, dependent 2-sample t-test). Patients with lower APACHE scores (0-15) have lower precision of 19.7% compared with 30.5% in the rest of the cohort (p-value 0.0004, dependent 2-sample t-test, Fig. 4b). Emergency admissions have lower recall of 88.2% compared with 93.6% for elective admissions (p-value 0.039), whereas surgical admissions have a higher recall of 92.3% compared with 87.7% in the rest of patients (p-value 0.039, Supplementary Fig. 9e).

***Model performance over time***

Fig. 5a shows how the performance of the model varies as time since admission increases. While the overall recall of the model is 90% (Supplementary Fig. 10 for other thresholds), the performance is not uniformly distributed across a typical patient's stay, with recalls of over 95% attained within the first 8 hours. After the first day, the overall recall of the model drops to around 83%. Supplementary Fig. 11a,b shows how event duration and time since the previous event affect recall. Using our dataset spanning 8 years, we also analyzed how changes in medical practice and patient characteristics may impact model performance in the future. We simulated the setting of using a trained model over a number of years into the future in a reverse way as follows: we fixed a test set comprised of patients admitted in 2016, and constructed eight training sets; one for patients admitted in each of the years 2008-2015. This design was chosen to eliminate the performance variability caused if one would use different test sets. In Fig. 5b, we report the performance of these eight models in terms of AUPRC and precision at fixed recalls (AUROC shown in Supplementary Fig. 11c). To control for the effect of training set size on performance, each training set was subsampled to have comparable size (2,366 patients). We observed a slight increase in performance the closer the test set is to the training set (Supplementary Fig. 11c). We observe different degrees performance variation the closer the test set is to the training set (Fig. 5b). By fitting an autoregressive model to the differences of the AUPRC values, we observe a first order term of size 0.14 which we interpret as the presence of a temporal drift[35]. This does not hold for the precision values (Fig. 5b), where we can assume stationarity (p-value $5*10^{-5}$, Dickley-Fuller test[35]).

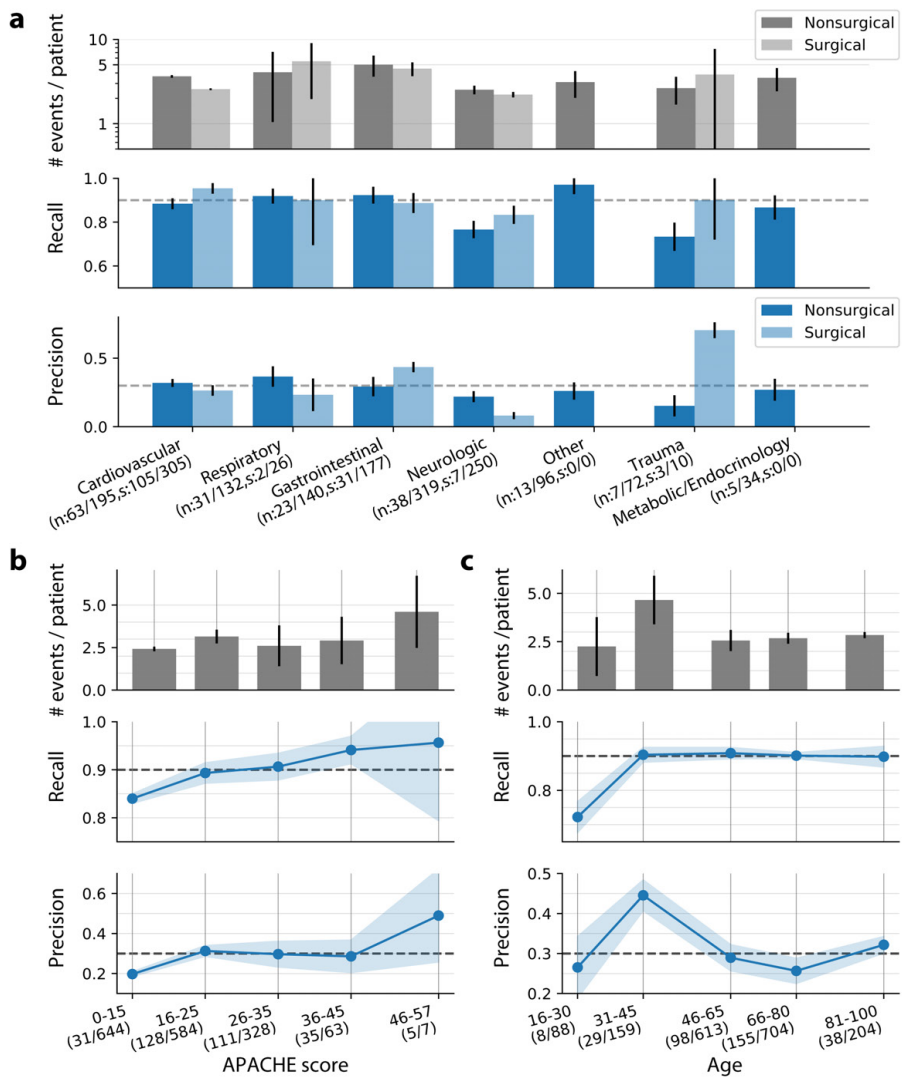

**Fig. 4: Performance in different patient cohorts.** Shown analyses use *circEWS* with a threshold corresponding to 90% recall, silencing of 30min, corresponding to an overall precision of 30%. **a:** Recall and precision for patients in different APACHE diagnostic groups. The notation (n: $a/c$, s:$b/d$) below each group name means that there are $a$ and $b$ patients with events among $c$ and $d$ patients in its non-surgical and surgical subgroup respectively. **b:** Recall and precision for patients stratified by APACHE-III score. The notation ($a/d$) under each group name means that there are $a$ patients with events and $d$ patients in the group. **c:** Recall and precision as a function of patient age.

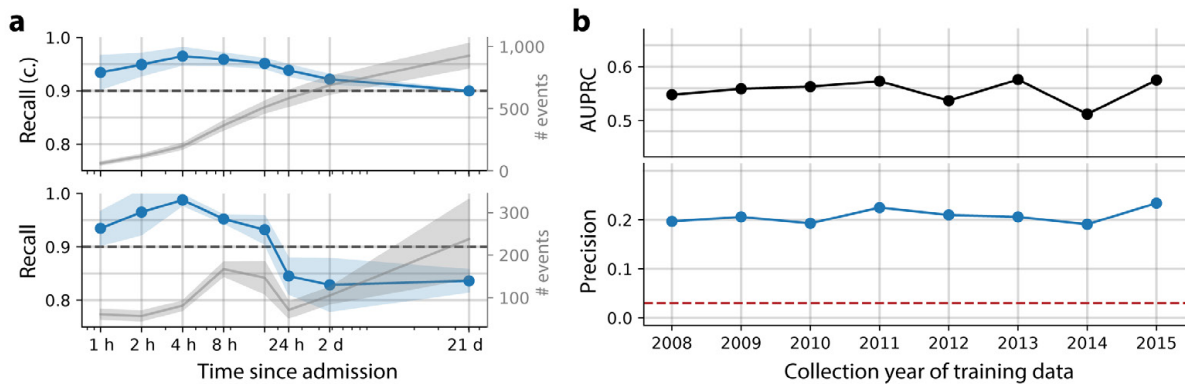

**Fig. 5 Performance over time. a:** Recall as a function of time since admission. Events (episodes of circulatory failure) are stratified based on time lag after ICU admission. In the top graph, we show the cumulative performance of the model, i.e., after 8 hours the overall recall of the model is approximately 96%. The lower graph reports the recall for each time period independently. **b:** Temporal generalization: we assessed how a model trained on historical data generalizes to a test set increasingly distant in the future. To do this, we fixed a test set using data from 2016 and train 8 models, each using one year of data between 2008-2016. We report here the AUPRC as a function of the year the model was trained, and the precision at a fixed threshold (baseline prevalence shown in red).

## Discussion

We have demonstrated that two variants of a machine learning-based early warning system (*circEWS* and *circEWS-lite*) can predict circulatory failure with very high recall – only a small fraction of events are missed and most events are detected several hours before an event. Since the prevalence of events is low, it is not easy to achieve a low false-alarm rate. Our system generates about one false warning per day and patient, which we consider very low compared to other warning systems in clinical practice and definitely low enough to be of clinical utility. *circEWS*, based on 500 features from 112 clinical variables performed best, but was only marginally better than *circEWS-lite,* based on 176 features from 17 of the most important variables. Performance was similar irrespective of diagnosis, severity of illness, and age – with a few notable exceptions (neurologic patients). Unsurprisingly, recall of the model was highest immediately prior to circulatory failure. Nevertheless, most events could be detected several hours in advance to allow for appropriate interventions to be initiated. The alarm system was tested on an independent patient group from a different hospital system and demonstrated very good performance, despite the lower temporal resolution of the available data.

The main limitations of our study are related to the single-center design, which creates a risk of over-fitting the model to the patient cohort and data at hand. However, the analyzed

ICU admissions originate from a population covering the whole spectrum of ICU patients and external validation demonstrates the applicability of our model in other ICUs. Further, it was not possible to retrospectively identify patients in whom supra-normal blood pressure values were targeted. High blood pressure targets are often set to maintain cerebral perfusion pressure in critically ill neurological or neurosurgical patients. These patients can have elevated lactate levels due to localized intracerebral ischemia or due to inadequate blood supply, or from surgical retraction[54–56], and therefore would fulfil our endpoint definition without being in circulatory shock. Their inclusion is likely to impair model performance, and we observed inferior performance of *circEWS* in this patient group.

*circEWS* was constructed with data collected in a clinical context, containing artefacts and errors. No manual data cleaning was performed. This ensures that similar performance can be expected once the model is applied on live data. The low prevalence of the endpoint was not artificially increased to improve apparent model performance, but left unchanged to mirror future applications of the model as realistically as possible.

Conventional systems that help identify patients at risk of circulatory deterioration in the ICU are based on variables known to determine circulatory function and tight alarm limits are set. Often lacking clinical relevance, such alarms can lead to alarm fatigue when medical staff is overwhelmed by excessive alerts[57]. In comparison to traditional alarm systems, *circEWS* integrates individual patient information and a large variety of physiological measurements from multiple organ systems to achieve a manageable number of timely alarms and thus reduces the chance of alarm fatigue[57]. While the reported precision of 30% at a recall of 90% appears low at first sight, the system raises an alarm only every 16h per patient and 82% of events are identified more than two hours in advance. This would give physicians enough time to evaluate the situation and take appropriate measures if needed. In our evaluation we are not able to detect cases where the physician intervened by treatment which lead to an improved state of the patient and to an avoidance of the deterioration event. These would be counted as false positives and decrease the estimated precision of *circEWS*.   Therefore, we expect that the precision in practice is higher than we estimate.

The ability to interrogate model predictions and to understand which features contributed to a given prediction is important for ensuring that this technology remains interpretable to its clinical users. Using *SHAP* values, we see that the model identified established predictors determined by circulatory state, but also time series representations, information from other organ systems, patient characteristics, and treatment parameters. Although in-depth analysis of feature importance is beyond the scope of this study, such an analysis in the future might offer new insights on previously unrecognized indicators of circulatory failure. Removing individual variables to inspect model performance indicated that of the top 20 variables identified by SHAP values, only the exclusion of lactate resulted in markedly decreased performance. This finding indicates that there is redundancy across variables and features; this effect is also seen when we ablate entire feature categories (Supplementary Fig. 12). The high relevance of the lactate variable suggest that more frequent measurements of lactate will likely lead to a performance increase and to an earlier detection of deterioration events.

To assess the external validity of *circEWS*, we applied it to the MIMIC-III ICU dataset. We observed that if we apply *circEWS-lite* directly to MIMIC, its performance remains high, degrading only slightly. Interestingly, when we retrain the classifier on MIMIC, the performance of the system only improves slightly, and for thresholds resulting in high recall, actually performs worse. This may be related to the quantity of data available in MIMIC-III to train models - the usable training data is much smaller (2.8 million vs. 13 million samples in HiRID) and the fact that the temporal resolution is much lower (in MIMIC, variables such as MAP are provided on average hourly - in contrast to a sampling frequency of 2 minutes in HiRID). As observed in Supplementary Fig. 5b, even with the large training size available in HiRID, the model's performance has not yet saturated with respect to training set size, and therefore, the more limited data in MIMIC may impair performance.

The practice of medicine changes as new research is conducted and new technologies are developed. Machine learning methods trained on historical data are therefore susceptible to reduced performance associated with future deployment[58]. Our results indicate a slight increase in model performance the closer the derivation is to the test set, providing evidence for this effect. Moreover, medical practice varies between providers as well as institutions. The importance of this locational dataset shift is illustrated in our results by the observation of better performance of the locally re-calibrated MIMIC retrain model versus the MIMIC validation. Our model should therefore not be seen as an unalterable and universal scoring system similar to traditional ICU scores. In a clinical setting, it will be important to continually monitor the quality of predictions using new data to constantly develop and re-calibrate the models to account for temporal changes and local differences in practice. The high time resolution and size of the HiRID data set make it valuable for further applications of machine learning. We are therefore in the process of making the data publicly available for research.

Machine learning techniques have been applied to tasks in radiology[59], pathology[60], and critical care[28,30] in retrospective clinical studies. Approaches spanning a spectrum of complexity have been developed to tackle clinical prediction problems, from linear models[61–63] to complex deep architectures[64]. In this work, we used gradient-boosted decision trees due to their observed superior performance in our application and ease of interrogation. This model class has been successfully applied in many different domains[65–67]. While we tested other models, including recurrent neural networks, we found these approaches inferior (see Supplementary Fig. 5e). This finding reflects recent observations[19] that careful feature design, combined with state-of-the art machine learning approaches, can outperform deep learning; in particular, when limited amounts of data are available. However, when more data is available for training the system it is likely that more expressive deep architectures may ultimately prove superior. Finding an existing system with which to compare *circEWS* is challenging. Methods have been developed to predict single aspects of circulatory failure including the time of initiation of vasopressors[66] or the onset of acute hypotensive episodes[67]. In the case of the former, an AUROC of 92% was achieved for vasopressor need within 2 hours, although other, clinically more relevant metrics are not reported. In the latter study, the algorithm predicted arterial hypotension 15 minutes before the hypotensive event with a sensitivity and specificity of 88% (AUROC 95%; precision was not provided). The prediction of septic shock has also been addressed, for instance, the TREWScore[22] identifies patients before the onset of septic shock with an AUROC of 83% (with 85% sensitivity at specificity of 67%). Ultimately none of these methods directly addresses our prediction problem, hence we have constructed an internal baseline method.

Our data shows that even short periods of circulatory failure are associated with an increase in ICU mortality. While this does not prove causality, it is safe to assume that preventing prolonged times of circulatory failure will likely decrease ICU mortality. Considering the demonstrated good performance of our models, we hypothesize that machine learning-based early warning systems may help ICU staff to more rapidly identify patients at risk for development of circulatory failure with a much lower false positive alarm rate than conventional threshold-based systems. Models developed on a local retrospective dataset can be transferred to other ICUs and be applied in the future. Nevertheless, performance will likely increase when models are retrained to the specific settings. Before clinical application of our models is possible, prospective research on the impact of model implementation on patient outcomes has to be conducted. Overall, we show that adaptive models have the potential to allow the shift from detection and treatment to automated prediction and hopefully prevention of organ system failure.

# Online methods

### Study design and setting

The study was designed as a retrospective cohort study for the development and validation of a clinical prediction model. The study was performed at the Department of Intensive Care Medicine of the Bern University Hospital, Switzerland (ICU), an interdisciplinary 60-bed unit admitting >6,500 patients per year. Data processing, model training, and analyses were performed at the Departments of Computer Science as well as Biosystems Science and Engineering at ETH Zürich, Switzerland.

### Ethical approval and patient consent

The institutional review board (IRB) of the Canton of Bern approved the study. The need for obtaining informed patient consent was waived due to the retrospective and observational nature of the study.

### Participants and data sources

The study included all patients admitted to the ICU in the period between the implementation of the ICU electronic patient data management system (PDMS; GE Centricity Critical Care, General Electrics, Helsinki, Finland) in April 2005 and August 2016. The PDMS was used to prospectively register patient health information, measurements of organ function parameters, results of laboratory analyses and treatment parameters from ICU admission to discharge.

The study flow chart is presented in Supplementary Fig. 2a. Patient admissions prior to 2008 were excluded from the analysis due to frequent changes in variable identifiers during the run-in phase of the PDMS implementation. Patients without data for determining circulatory failure and patients receiving any form of full mechanical circulatory support, younger than 16 years or older than 100 years, or actively declining the use of their data for research purposes were excluded.

### Analysis platform

All computational analyses were performed on a secure compute cluster environment located at ETH Zürich (https://scicomp.ethz.ch/wiki/Leonhard). Python3, with numpy[68], pandas[69] and scikit-learn[70] formed the backbone of the data processing pipeline.

***Artefact removal***

Artefact removal and correction was performed using variable-specific algorithms to enable future live deployment and constituted a major effort. Four main types of artefacts were identified:

- *Timestamp artefacts:* Measurement time information was stored in two fields - the time the measurement was taken (*SampleTime*), and the time it entered the system (*EnterTime*). While the latter field was automatically filled, *SampleTime* can contain manual input errors, such as an incorrect month or even year, disrupting the order of the time series, or falsely indicating unreasonably long ICU stays or gaps between measurements. Intervals longer than 1 day were identified and corrected as described in Supplementary Tab. 3. Timestamp artefacts existed in 3,530 (8%) of patient admissions.

- *Variable-specific artefacts:* Blood gas samples required a manual selection of the sample type as arterial or venous. As arterial is the default option, multiple venous samples were wrongly labeled as arterial. This was identified by comparing the oxygen saturation in the blood gas sample to the central venous saturation - if this difference was less than 10% of the standard deviation of oxygen saturation (across the training data), the sample was re-labelled as venous. Patient height and weight were manually entered and sometimes accidentally interchanged. For weight/height measurements resulting in Body Mass Indices (BMI) > 60kg/m$^2$ or < 10kg/m$^2$, height and weight were swapped if this resulted in a BMI in the range of 10-60 kg/m$^2$.

- *Out of range artefacts:* For each variable, a range of possible values (including pathologic values) was defined - these are reported in Supplementary Tab. 5 (column "*permitted range*" in *variables* tab). Values outside this range were deleted.

- *Record duplication:* The database contained records of the same variable for the same patient with the same timestamp. For n*on-pharmaceutical variables* one value was kept if the values of the duplicates were identical. Otherwise, we compared the standard deviation of the duplicates with the global standard deviation of the variable across all patients. If the former was <5% of the latter, we kept the mean of the duplicates. Otherwise, the duplicates were considered unreliable and deleted.

- For *pharmaceutical variables* duplicates with an entry indicating a "zero" dose were deleted. For duplicates of drugs applied as tablets or injections the sum of the recorded dose values was kept. For duplications with a status indicating none of the above, we took the mean of the dose.

- *Processing pharmaceutical variables:* We converted all pharmaceutical variables to either a rate or presence indicator. Drugs given as boluses such as injections and tablets were converted to an effective continuous rate over a time-period specified according to the estimated duration of action (Supplementary Tab. 5, column "acting period (individual)" in *drugs* tab). In cases where a quantitative rate is not possible, we used a binary flag to indicate if the drug (or drug class; see next section) was present.

### *Variable merging*

The PDMS contained many instances of the same parameter being recorded using different identifiers (e.g., different dilutions of vasopressors, different probe locations for core temperature measurements). Moreover, specific variables were infrequently observed (e.g., foscavir was observed < 50 times), but belonged to a meta-category (such as "Antiviral therapy"). To build a model which is less specific to the local patient cohort/setting, we used the following variable merging strategies, reducing our set of variables from 710 to 209. Identical medical core concepts recorded as different variable ids were merged (e.g., different probe locations for core temperature measurements). Identical pharmaceutical compounds were aggregated into one variable (Supplementary Fig. 4a).

Certain clinically less important compounds were aggregated to group variables regarding the targeted pharmaceutical effect (e.g., non-opioid analgesics, Supplementary Tab. 5, columns "*drug*" and "*constituent drugs (if relevant)*" in *drugs* tab). This was performed for better temporal and inter ICU generalizability by making the model features independent of the specific compound used. If this led to multiple measurements at the same time, the following strategies were used. For physiological parameters (such as temperature) or lab tests, we used the median of simultaneous measurements. For pharmaceutical variables, we used a weighted sum over simultaneous infusions, with weighting given by effective relative doses determined by analysis of the literature. Otherwise, we merged variables into a binary indicator denoting whether or not any drug from that class (e.g., antibiotics) was present, or count how many drugs are present (Supplementary Tab. 5, column "merging ratio" in *drugs* tab).

### Circulatory state annotation

We annotated every 5-minute interval of a patient's stay with their current circulatory state using three types of variables: lactate (arterial and venous), mean arterial blood pressure (MAP), and presence of vasoactive/inotropic drugs. The state was established using a window of 45 minutes duration centered on the current time-point. To reduce spurious calls due to transient states, in each such window all conditions had to be independently true for 30 minutes (not necessarily consecutive).

We defined the following three states:

- Patient currently not in circulatory failure: if MAP is >65 mmHg, vasoactive/inotropic drugs are not present, and lactate is ≤2 mmol/L.

- Patient currently in circulatory failure: MAP is ≤65 mmHg or (not exclusive) vasoactive/inotropic drugs are present and lactate is >2 mmol/L

- Unknown/Ambiguous: if any of the following conditions hold:
  - No MAP or (interpolated) lactate is available in the 45-minute window
  - MAP or vasoactive/inotropic drug criterion is met, but lactate is ≤2 mmol/L

To enable state annotation at all time-points, we imputed lactate values between measurements. We linearly interpolated lactate values between measurements, unless the patient's lactate value had passed the threshold of 2 mmol/L in either direction. If a patient's state had changed, from either low to high lactate or vice-versa, we linearly interpolated depending the interval between the two measurements. If they were less than six hours apart we interpolate for the full period. Otherwise we forward/backward filled for a maximum of 3 hours and the remaining time points were left missing.

To handle the starts/ends of the stay, we filled forward/backward. If the patient's first/last measurement was "normal" (under the threshold), we backward/forward filled indefinitely. If the measurement was abnormal, we filled backward/forward for up to 3 hours.

As this imputation scheme implicitly used information from the future, it was only used for annotating (and subsequently labeling) timepoints. Adaptive imputation and feature generation for model development were performed independently and as described below without using future information.

***Patient-centered adaptive time series imputation***

Imputation parameters were pre-computed on the training set. They consist of the median/IQR of the sampling interval of each non-medication variable $i$, denoted as ($m_i$, $iqr_i$) below. The imputation process created a time grid with step size 5 minutes, starting and ending at the patients first and last heart rate measurements, or ending at 28 days after admission (whichever is shorter). This provided a unified definition of "beginning of stay", corresponding to the start of basic monitoring, and avoids biasing the data towards patients with very long stays. Values were imputed for all variables independently at each grid-point using the following process. Prior to the first measurement of a given variable, or if the patient had no measurements, we filled it in using a normal value (Supplementary Tab. 5, column "default value" in *variables* tab). If the last measurement, as seen from the grid point, was less than $m_i$ +$iqr_i$ minutes away, we used forward filling from the last value. Otherwise we linearly returned to the median of the last 2*($m_i$ + 2*$iqr_i$) minutes, as measured from the point where we entered the region where this imputation mode is applied, for 2*($m_i$ + 2*$iqr_i$) minutes in total. After that, we assumed that the value stayed constant at this median value (indefinite forward-filling), until the next valid measurement, if any, at which we returned to step 2. Static variables were imputed according to either the mean or the mode value in the training data, for continuous and categorical values respectively.

***Feature generation***

Feature generation took as input the imputed data and generated features sample-wise on the 5-minute grid. The first 30 minutes of a stay were ignored for feature generation, because the history of vital signs and lab tests contained insufficient information to generate reliable features. Six types of features were generated for each time grid point. They included the current estimated value of a variable, and five others, which are described in detail below. Besides these feature classes, we also added "time since admission" as an individual feature.

- *Static features:* Six static features (Age, Indicator of Surgical Admission, Indicator of Emergency Admission, APACHE Patient Group, Height and Sex) were concatenated to each time-grid sample of a patient.

- *Multi-resolution summaries:* To capture the temporal history of our data, we constructed time windows of increasing size and extract summary statistics over each window (Fig. 1d). The window sizes and statistics depended on the variable type and sampling rate. We classified each variable as either of high, low, or medium frequency according to its median sampling interval in the training set, or estimated duration of action for medications (Supplementary Tab. 5, in *drugs* tab). For each frequency category, we defined four time windows for short, medium, long, and very long time horizons using prior clinical knowledge (Supplementary Fig. 13a). For continuous-valued variables, we extracted the median, IQR, minimum/maximum values, and a trend estimate for each of the time horizons. We reported a mean estimate rather than median for medications. For categorical and binary variables, we reported only the mode or mean respectively. Additionally, we summarized the entire stay up to now with the summary functions mode, mean, or median depending on the variable category.

- *Instability history features:* Assuming that patients who have already suffered circulatory instability are at increased risk of recurrence, we formed a set of features to capture the patient's history of instability. We encoded the current state, time to the last pathologic state as well as the density of pathological states in the past. All of these refer to logical sub-conditions of our circulatory failure definitions (Supplementary Tab. 13b). If no abnormal state was measured, we set a symbolic value of 30 days (larger than the maximum length of a stay). The density was defined as the ratio of the duration the state was active and the length of the stay so far.

- *Measurement-intensity based features:* Since imputation removed information about when and how often measurements were performed, we reintroduced some of this information with this feature class for vital sign measurements and lab tests. We report the time since the last (non-imputed) measurement (30 days, if no measurement available) as well as the ratio of time points with measurements in the stay up to now.

- *Shapelet-based features:* A shapelet is a small time series subsequence that is discriminative for the class label and known to capture salient temporal dynamics of time series in a variety of application domains[71,72]. We used the computationally efficient S3M method[73]: For each variable, 300 subsequences were extracted with a padding of 5 min before any deterioration event; these were labelled as cases (Supplementary Fig. 13c). The same number of uniformly sampled time series from the remaining patients served as controls. The remaining parameters were adjusted according to the resolution of the variable (see Supplementary Fig. 13c). As the resulting shapelet set might be very large (up to multiple 1,000 shapelets per variable), the subsequent shapelet selection step created a feasible number of shapelet features per variable (in our case 20 shapelets per variable and length) that was representative of the space of all shapelets with the min-max approach: First, the shapelet with the highest accuracy in differentiating cases from controls on the training data set was selected. Afterwards, shapelets were iteratively selected such that the minimal distance to the set of already selected shapelets was maximized until 20 shapelets are chosen (Supplementary Fig. 13d). Supplementary Fig. 12d shows that the min-max sampling gave similar performance as random sampling or selecting the top 20 shapelets. A shapelet was used to construct a set of features per time point by concatenating the L2 distances between the shapelet and the history of a patient's variable during the last 4 hours. This history of distances (dist-hist) approach outperformed other feature computation approaches as shown in Supplementary Fig. 12d: Single distance (distance), the minimum over all distances in the last 4h (min) and counting the number of shapelets that have occurred in the last 4h (count).

### *Supervised learning of deterioration prediction*

We defined a binary prediction task to be performed every 5 minutes while a patient is not in circulatory failure (otherwise, a warning would be too late). At each such time point, the model predicted whether the patient will develop circulatory failure in the next 8 hours (positive label), or not (negative label). If the patient's circulatory state during the next 8 hours cannot be determined, no label was assigned and the time-point was excluded from training and evaluation. To avoid overestimating the prospective performance of our model (see Supplementary Fig. 5c for an illustration of this effect), we used an experimental design in which the test set contains the most recently admitted patients in the cohort (see Supplementary Fig. 4b); we call this a "temporal split". In a given temporal split, the full methodology was applied independently, including missing data imputation, feature extraction, model training and hyperparameter selection.

Five overlapping temporal splits were constructed, each containing admissions across 5 years (Supplementary Fig. 4b). The admissions of the last year were split 1:1 to define the validation and test sets, respectively. The remaining earlier admissions were used as training set of that split. The start of each subsequent split was shifted by six months. Further, the most recent 10% of patient admissions (November 2015 to May 2016) were defined as the held-out evaluation set which was not used for model development to avoid subtle overfitting to this data-set. This set was also split 1:1 into held-out validation and test sets. Using the rest of the available data as training data, this held-out set formed a special "held-out" split, which was used to provide a point estimate of model performance. The five temporal splits were used to estimate the variability of model performance, containing disjoint test sets, and partially disjoint training sets. We report this variability as the standard deviation over model performance in these splits. Lastly, an exploration split was defined which assigns patients at random to training, validation and test set in proportions 8:1:1. This split was used to compare temporal generalization to the standard approach of randomly assigning admissions to training and test set (Supplementary Fig. 5c).

***Machine learning approaches***

*We compared the following three state-of-the-art supervised machine learning techniques to learn to detect deterioration events:*

- *Gradient boosted ensemble of decision trees, and decision tree baseline:* The gradient boosting library lightGBM (version 2.2.1) was used for model fitting[33]. The hyperparameter settings maximizing the AUPRC on the validation set were used to generate the predictions on the test set, after refitting on the training set. The model training process was stopped if the AUPRC on the validation set did not improve over 50 consecutive fitting iterations, resetting the model state to the best iteration before early stopping. Since lightGBM can deal natively with categorical data, we did not one-hot-code such data before model fitting. As this model achieved highest performance during system development, it was used for further analyses. To obtain the decision tree baseline, we set the number of trees to 1.

- *Logistic regression:* The class SGDClassifier from the scikit-learn library (version 0.20.0) was used for model fitting[70]. The strength of the regularization parameter was selected by maximizing the AUPRC on the validation set. Before fitting, continuous features were standardized (zero mean, standard deviation one), and categorical features one-hot-encoded.

- *LSTM-based recurrent neural network model:* We constructed a long short-term memory (LSTM)[74] network comparison in TensorFlow 1.11.0[75]. We used the same set of features as provided to lightGBM for fair comparison after intermediate results suggested worse performance when using only the raw variable values. Since a fraction of the features are static, a small-size single-layer-perceptron (SLP) was used alongside the LSTM to learn from the static features. The LSTM and the SLP output the hidden states for the dynamic and the static features respectively, and by linear combination these two hidden states are fed into the output layer. Before training, all non-categorical features were standardized and categorical features one-hot-encoded.

Hyperparameter settings and grids for the following models are listed in Supplementary Tab. 2, if not otherwise described.

### *Variable and feature selection*

The importance of individual features was measured using mean absolute SHAP values of predictions made on the validation set for each temporal split. Before SHAP values were computed, the negative instances in the validation set were sub-sampled to achieve a balanced dataset. The variable ranking was obtained with a greedy forward selection approach where the variable associated with the feature with the largest mean absolute SHAP value considered the most important variable. This variable and all its features were then removed from the ranking and the procedure was repeated. The final ranking of important clinical variables was determined using the held-out split. The standard deviation of the ranks is computed on the five temporal splits used for model development. Optimal model performance was obtained using 500 features, and removing more features degraded performance (Supplementary Fig. 12e). These features, comprising 112 variables, are provided to the full model. We further identified the top 20 clinical variables using the ranking procedure (see Table 1) and excluded three variables not identifiable in MIMIC ("non-opioid analgesics" and two inotropes). The resulting 176 features from the remaining 17 variables formed the compact model.

### Early warning system and evaluation

A core contribution of this work is an early warning system for circulatory failure within 8 hours - *circEWS*. We built two variants, *circEWS* and *circEWS-lite*, based on the binary classifiers "full" and "compact" described above. The output of the classifier is a score between 0 and 1, which is converted to an alarm if it exceeds a fixed threshold. On top of this, we employed a silencing policy to reduce unnecessary repetitive alarms: For 30 minutes after an alarm is raised, any potential subsequent alarms were suppressed. If a patient experienced circulatory failure and recovered during this silencing period, the system was reset to allow new alarms after 25 minutes. Effects of different silencing periods and system reset times are shown in Supplementary Fig. 6. Our objective was to evaluate *circEWS* in a clinically relevant context, focusing on the percentage of circulatory failure events the system is able to detect and the rate of false alarms. Model precision was defined as the fraction of alarms that correctly predict the onset of an event (a period of circulatory failure) within the next 8 hours. Model recall was defined as the fraction of events that are captured by an alarm. This is analogous to exon prediction in gene finding[76]. Significance of performance differences in patient sub-cohorts was assessed using a $p<0.05$ cutoff using dependent 2-sample t-tests corrected for multiple-testing with the Benjamini-Hochberg procedure, matched on the 6 temporal splits in which the experiment was replicated.

### External validation on MIMIC-III

MIMIC-III v1.4 was used for external validation, including only patients admitted after 2008 and the introduction of MetaVision. 17 of the 20 most important variables (Table 1) were identified and extracted. Non-opioid analgesics could not be matched. The drugs levosimendan and theophylline were not used at Beth Israel Deaconess Center and were excluded. Since many of the artefact removal steps described above are specific to HiRID, we applied only artefact removal using the same fixed variable specific ranges on the MIMIC data. MIMIC data was converted into the correct format to be processed by the rest of the HiRID pipeline. Patient state annotation, label generation, and missing data imputation were performed as described above with minor modifications.

*Imputation parameters:* Part of our imputation pipeline required calculating the sampling interval for each variable. These intervals were not recomputed to provide similar data for validation to our existing model. Furthermore, we did not expect the ground truth of these values to vary much between ICUs, even if different down-sampling is used.

*Time grid:* The temporal resolution of MIMIC was different to that of HiRID. Nonetheless, to mirror the HiRID data structure as closely as possible, we resample MIMIC to a five-minute grid, even if this introduces a large quantity of imputed data.

We consider two settings for evaluating *circEWS-lite* on MIMIC: "MIMIC (Validation)", and "MIMIC (Retrain)". In the validation setting, we applied *circEWS-lite* on MIMIC, using the full dataset as a test-set (in total 9,040 patients; see Supplementary Fig. 2b). In the retrain setting, we applied the same processing as before, but trained a lightGBM classifier to predict circulatory failure on MIMIC. We formed five "replicates" of MIMIC using Monte Carlo resampling, in each replicate assigning admissions at random to training/validation/test sets in the ratio 3:1:1. Since absolute admission times were not available in MIMIC, temporal splits could not be constructed. Our final performance estimate was the mean of the performance in each replicate's test set, and the error estimate was the standard deviation over replicates. In the retrain setting, the final size of each training set contained approximately 7,950 patients, with ~1,000 patients in the test sets.

**Prevalence correction for MIMIC**

The test sets used for MIMIC (validation) and MIMIC (retrain) have different prevalences of positive labels/events compared to the test set of the HiRID dataset. Therefore, to enable a comparison of the performance of *circEWS-lite* in terms of precision and recall between HiRID and MIMIC, we calibrated the precision-recall curves for MIMIC in Fig. 2b,d and Supplementary Fig. 7b such that MIMIC would have the same positive label/event prevalence as HiRID. The uncorrected precision-recall curves are shown in Supplementary Fig. 7a,c. The calibration was performed in different ways for the time point-based and alarm/event-based precision recall curves:

- In the time slice-based evaluation the number of false alarms observed in the MIMIC test sets was down-scaled by multiplication with

$$s = \left( \frac{1}{\text{prev}_l(\text{HiRID})} - 1 \right) \Big/ \left( \frac{1}{\text{prev}_l(\text{MIMIC})} - 1 \right)$$

where $\mathrm{prev}_l(d)$ refers to the positive (time-slice) label prevalence in the dataset $d$.

- For the calibration of the alarm/event-based precision-recall curves for MIMIC, we computed the AUPRC of an alarm system that was based on a random classifier and with the same silencing policy as *circEWS-lite* for both HiRID and MIMIC, which we denote by $\mathrm{prev}_e(\mathrm{HiRID})$ and $\mathrm{prev}_e(\mathrm{MIMIC})$, respectively. The AUPRC of the random alarm system is the event prevalence of the dataset. We downscale the number of false alarms observed in the MIMIC test set with:

$$s = \left(\frac{1}{\mathrm{prev}_e(\mathrm{HiRID})} - 1\right) \Big/ \left(\frac{1}{\mathrm{prev}_e(\mathrm{MIMIC})} - 1\right)$$

to satisfy the assumption that the calibrated MIMIC dataset has the same event prevalence as HiRID.

The correction factor $s$ is then multiplied with the false alarm counts when computing precision on the MIMIC data in order to obtain corrected precision estimates.

### *Data and code availability*

The research code used to perform experiments reported in this work will be published under an open source license. We will also release the processed data for future studies under a controlled access mechanism upon publication. We are currently discussing the details of sharing the details with Physionet[78].

## Acknowledgements

Funding from this work was provided by the Swiss National Science Foundation (grant #176005 to G.R. and T.M.M.). G.R. and S.L.H. received core funding from ETH Zürich. We gratefully acknowledge helpful discussions with Heiko Strathmann and Viktor Gal. We thank Victoria Andreas, Ximena Bonilla, David Sidebotham, Nora Toussaint and Irene Jarchum for proofreading the manuscript.

## Author Contributions

S.L.H., M.H., X.L., M.F., G.R., T.M.M., K.B., T.G. designed the experiments; M.F., T.M.M. selected and provided the clinical data and context; X.L., M.F., S.L.H, M.H. with contributions from T.M.M., M.Z. and G.R. preprocessed and cleaned the data; S.L.H., M.F., T.M.M. with contributions from G.R., X.L., M.H. defined and developed the labeling of deterioration events; M.H., M.F., with contributions from X.L., S.L.H., T.M.M., G.R. devised and implemented the adaptive imputation strategy; M.H., M.F., X.L., S.L.H. developed and extracted non-shapelet features, T.G. and C.B. developed code for shapelet analysis. M.H., X.L., S.L.H. developed the pipeline for supervised learning; X.L. implemented the LSTM model; T.G. implemented the decision tree baseline. C.E., C.B., M.Ho., M.M., B.R., D.B. contributed to various analyses of the data; T.M.M., G.R., S.L.H., M.F. and K.B. conceived and directed the project; S.L.H., M.F., M.H., X.L., T.G., T.M.M., G.R., and K.B. wrote the manuscript with the assistance and feedback of all the other co-authors. S.L.H. and T.G. with input from all authors created Fig. 1.

## Competing Interests

The authors declare no competing interests.

# Machine learning for early prediction of circulatory failure in the intensive care unit

Stephanie L. Hyland[1,2,3,4,+], Martin Faltys[5,+], Matthias Hüser[1,4,+], Xinrui Lyu[1,4,+], Thomas Gumbsch[6,8,+], Cristóbal Esteban[1,4], Christian Bock[6,8], Max Horn[6,8], Michael Moor[6,8], Bastian Rieck[6,8], Marc Zimmermann[1], Dean Bodenham[6,8], Karsten Borgwardt[6,8,‡], Gunnar Rätsch[1,2,3,4,7,8,‡], Tobias M. Merz[5,9,‡]

[+] Joint-first authors
[‡] These authors jointly directed this work. Contact: tobiasm@adhb.govt.nz, gunnar.ratsch@ratschlab.org, karsten.borgwardt@bsse.ethz.ch

## Electronic supplementary information

- **Supplementary Figures 1-13**

- **Supplementary Tables 1-5**

**Supplementary Figure 1**

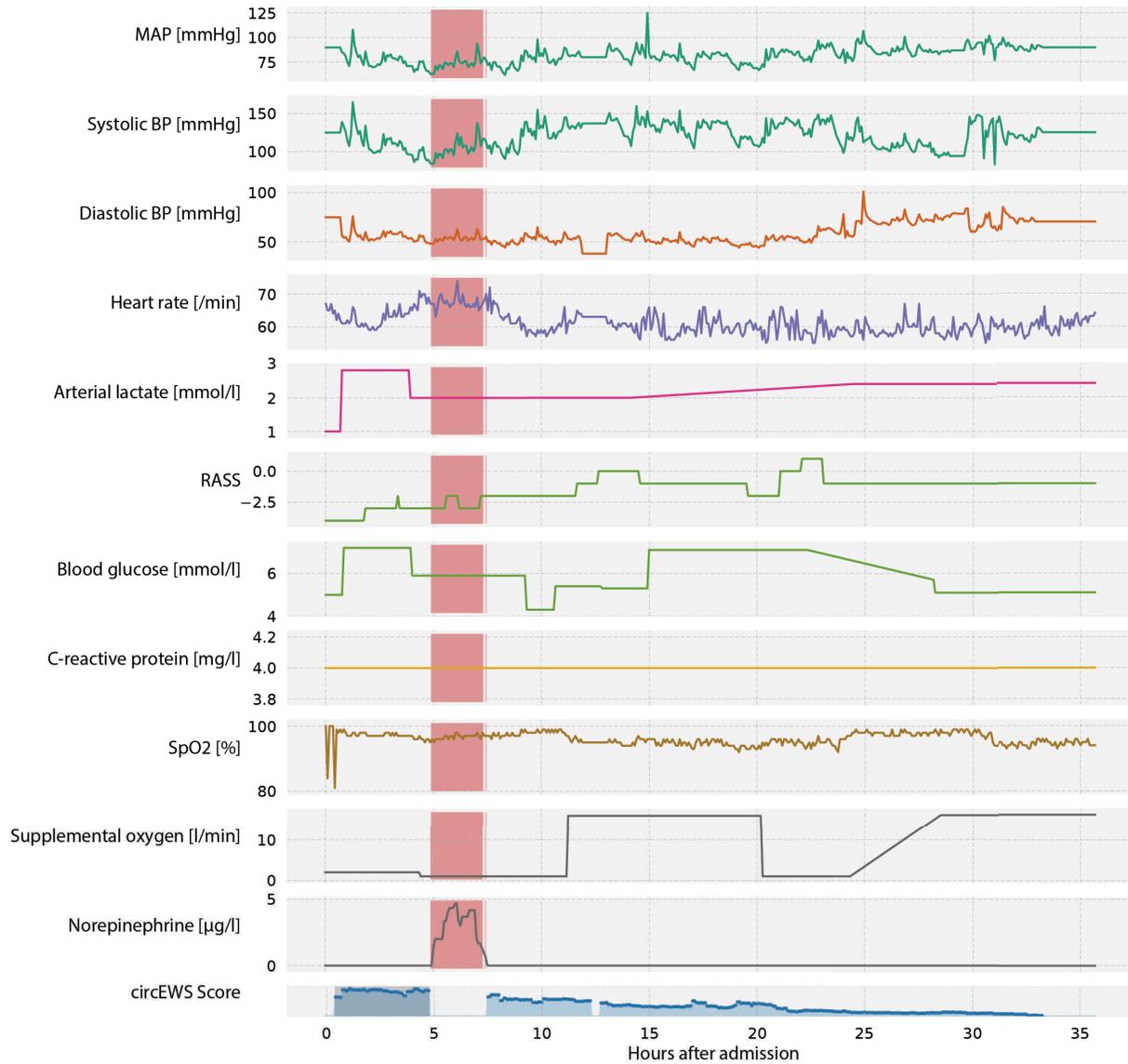

**Supplementary Figure 1: Example patient stay.** Visualization of an example patient stay with time after admission on the x-axis. Shown are the recorded values for this patient for the top-ranking variables (Table 1). The bottom time series shows the prediction score of *circEWS*. The red region denotes an area where the patient is in circulatory failure. The dark grey area overlaid on the score time series (bottom) denotes the region where the event should be predicted (8 hours before the beginning of circulatory failure). RASS: Richmond Agitation Sedation-Scale.

**Supplementary Figure 2**

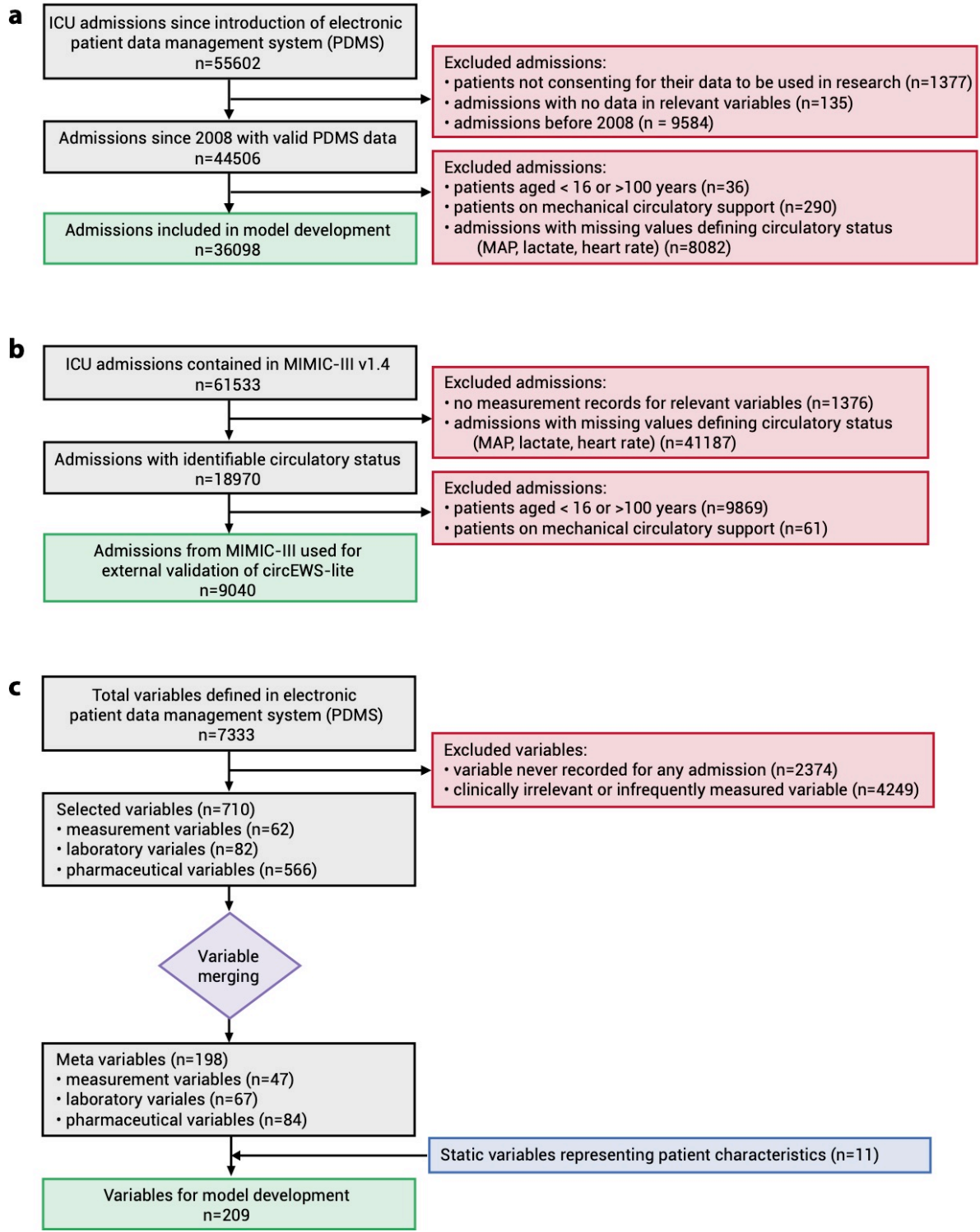

**Supplementary Figure 2: a:** Flow chart of the exclusion criteria applied to the HiRID base cohort. **b:** Flow chart of the exclusion criteria applied to the MIMIC-III cohort. **c:** Flow chart of the exclusion, merging, and post-processing applied to the variables in the patient data management system of the HiRID cohort.

**Supplementary Figure 3**

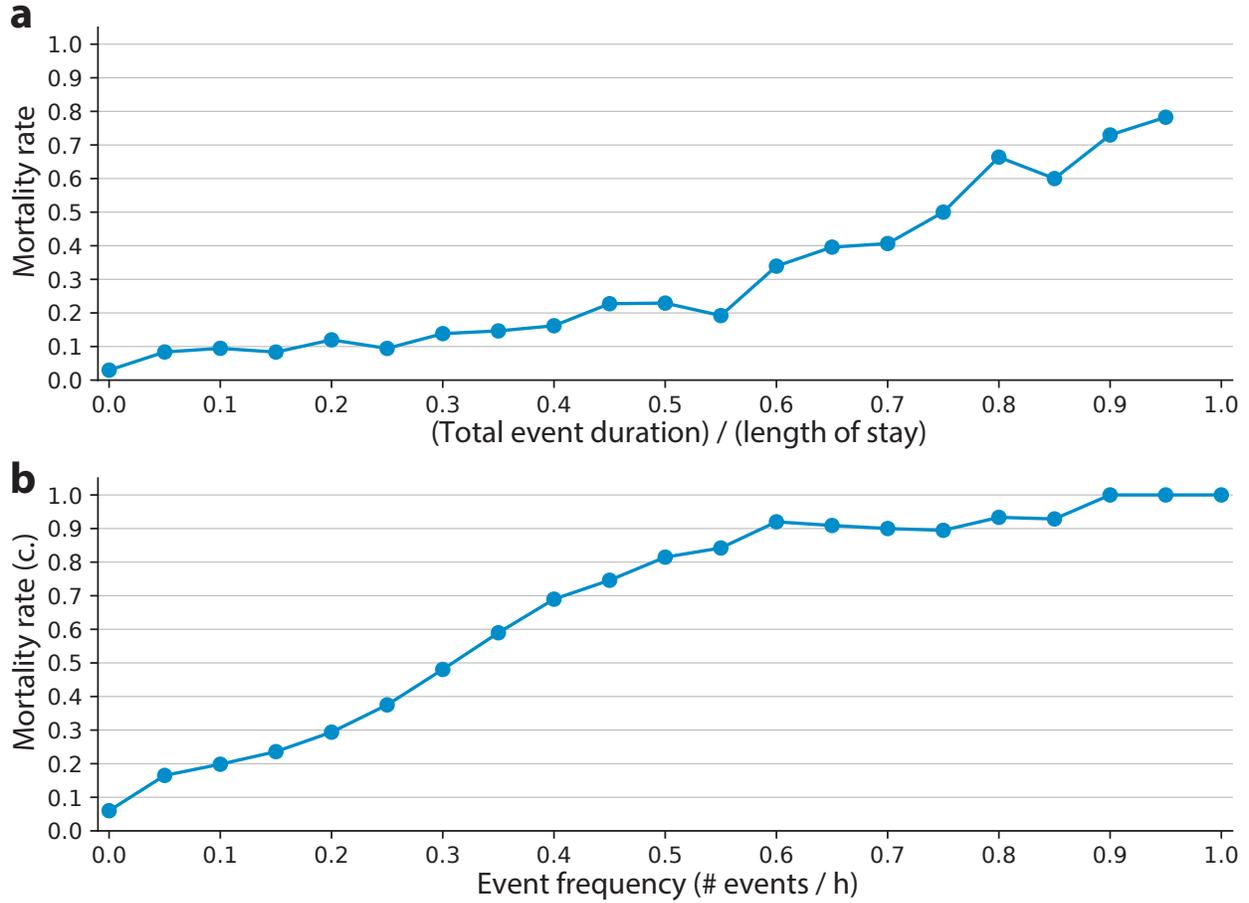

**Supplementary Figure 3: Correlation of duration of circulatory failure and mortality. a:** Shown is the mortality rate as a function of the duration of circulatory failure expressed as a fraction of length of stay in ICU. **b:** Shown is the *cumulative* mortality rate as a function of the frequency of occurrence of circulatory failure per hour of ICU admission.

**Supplementary Figure 4**

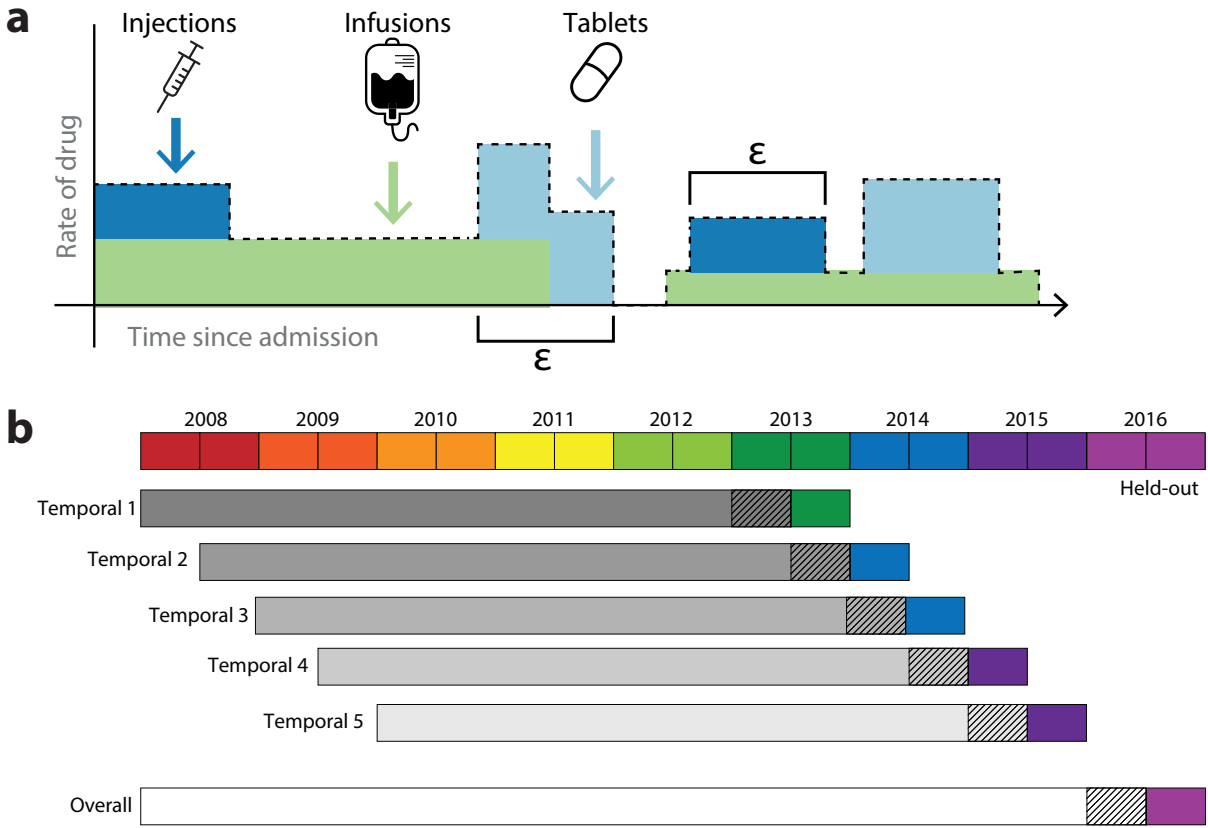

**Supplementary Figure 4: Data processing and experimental design. a: Processing details for pharmaceutical variables.** An effective flow rate of a drug was calculated by merging continuous infusions, intravenous boluses, and tablets. Intravenous bolus applications and tablets do not result in a constant effective blood drug level and were therefore converted to an effective rate by dividing the given dose by the acting period of the drug. These acting periods were specified using clinical domain knowledge and are listed in Supplementary Table 5b. **b: Experimental design.** We formed five "replicate" splits with disjoint test sets (coloured, dark), and partially overlapping training sets (grey). The validation and test sets consisted of the most recently admitted patients, where the validation patients (hatched) are from an earlier period. This enabled us to select a model more likely to be generalizable to the future. In addition, a held-out split was formed. The test set of the split was not used for developing the models described in the manuscript and only assessed for preparing the publication figures to avoid subtle overfitting to the data set.

**Supplementary Figure 5**

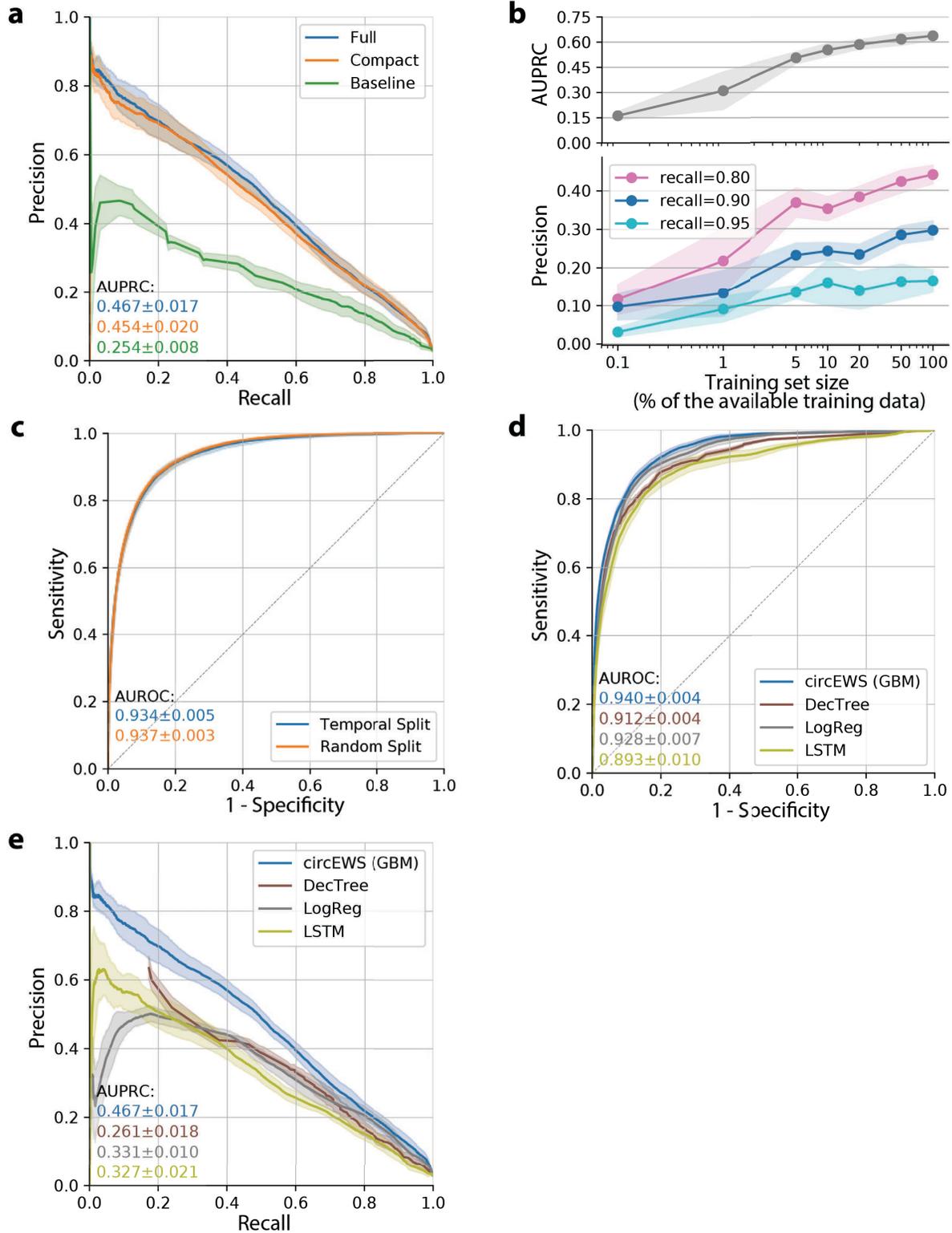

**Supplementary Figure 5: Model & Training.** Evaluation using the classical way of defining recall/precision based on correctly predicted labels. (Panels **a,c,d,e**). **a**: **Time slice PR on HiRID.** Precision-recall curve comparing the full, compact, and baseline models on the task of predicting circulatory failure. **b**: **Effect of training set size.** Analysis of impact of training set size to address the question whether our model performance has "saturated" on the dataset size. This was done by artificially subsampling patients at random and retraining the model. This analysis was performed using the *circEWS* evaluation policy. We observed that model performance decreases drastically beyond sub-sampling to less than 5% of the original training set size, and the model did not show obvious saturation effects as we move to the full size of the data. **c: Temporal vs. random data split.** Comparison of prediction model performance using a random data split into train, validation and test sets, irrespective of patient admission times, and the temporal split strategy used in all other analyses. As the prevalences of the test sets were different, negatives samples of the higher-prevalence test set were sub-sampled. We observe that the unrealistic evaluation setting of using a random split over-estimates performance compared to retrospective model construction. **d-e: Comparison of machine learning models.** Comparison of machine learning approaches using ROC/PR curves. We considered a simple linear baseline (logistic regression; LogReg), a tree-ensemble based method (based on lightGBM, "GBM"; used to construct *circEWS*), an individual decision tree (based on lightGBM, "Tree"), and a recurrent neural network ("LSTM"). The Tree models received identical input as given to GBM. The LogReg and LSTM received normalized feature values. We observed that gradient-boosting ensembles clearly outperform the other methods, followed by LogReg and LSTM/Tree models.

**Supplementary Figure 6**

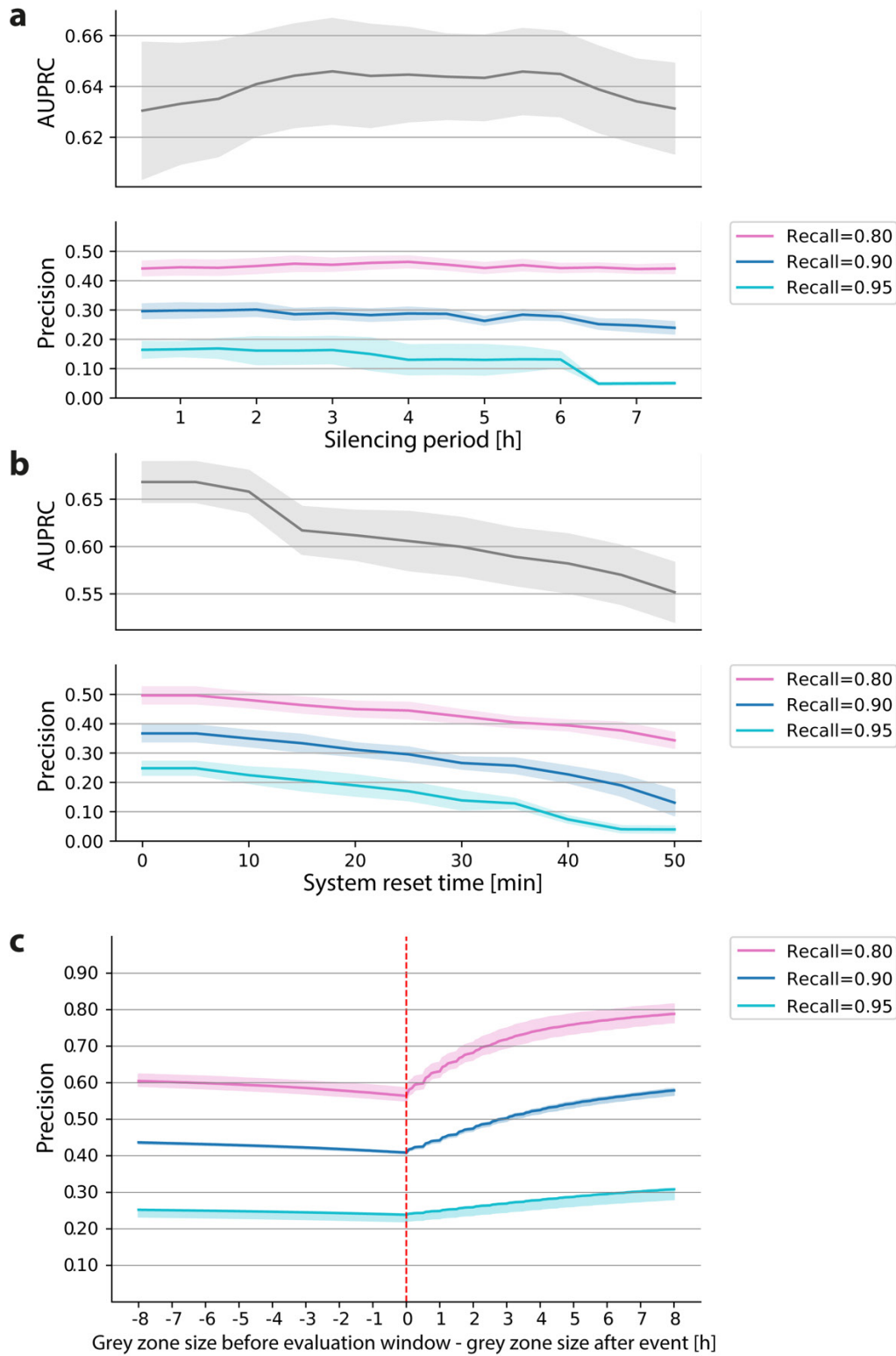

**Supplementary Figure 6: Silencing & resetting policy. a: Effect of silencing period.** We fixed a silencing window of 30 minutes. Here we show how overall performance (assessed by AUPRC and precision at fixed recall) varies as we modify this silencing time. We see that up to 7 hours of silencing results in diminished, but still high performance.

The long silencing time is mitigated by the unsilencing mechanic, whereby a patient entering and recovering from a period of circulatory failure will cause silencing to be terminated early. **b: Reset time.** We study the impact of silencing the alarm after a patient exited from a state of circulatory failure, the reset time. Increasing the reset time results in lower AUPRC and precision due to missed deterioration events. The default value for reset time in *circEWS* is 22.5 minutes. **c: Grey zone.** We studied the effect of a grey zone where alarms are not counted as false positives. A grey zone of one hour did not lead to significantly larger precision in any direction at a p-value threshold of 0.05 using a student two-sample *t*-test.

**Supplementary Figure 7**

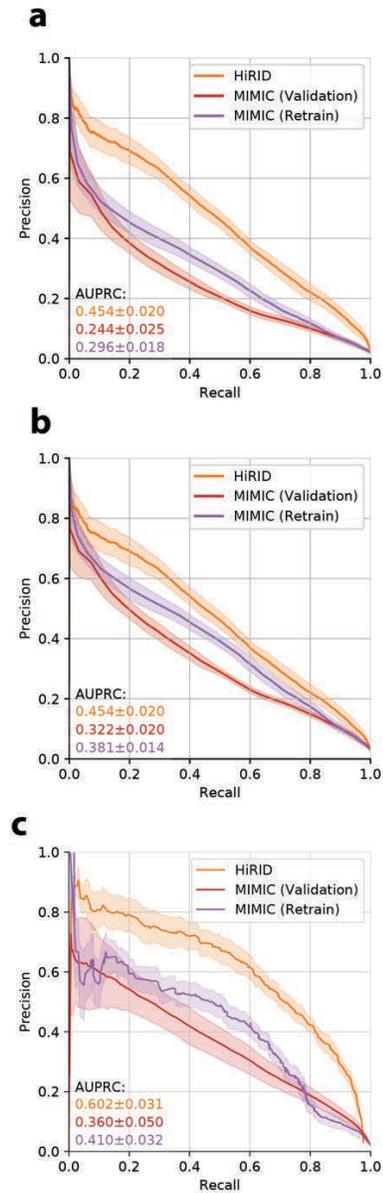

**Supplementary Figure 7: Evaluation of external model validation. a: Time-slice PR (uncorrected).** Time-slice based PR curve evaluated in the classical way with respect to correct predictions of the label, comparing the performance of the compact model on the HiRID data-set and the external validation data-set (MIMIC). The curves are not corrected to account for the lower positive label prevalence in the MIMIC dataset. **b: Time-slice PR (corrected):** The same experiment of panel **a** but correcting the MIMIC curves for the substantially lower label prevalence in the MIMIC data-set, implying equal performance of random classifiers on both data-sets. **c: Alarm system evaluation (uncorrected).** Evaluation of the *circEWS-lite* alarm system using an alarm/event-based recall metric, uncorrected for the prevalence differences of the MIMIC and HiRID data-sets, as in panel **a.**

**Supplementary Figure 8**

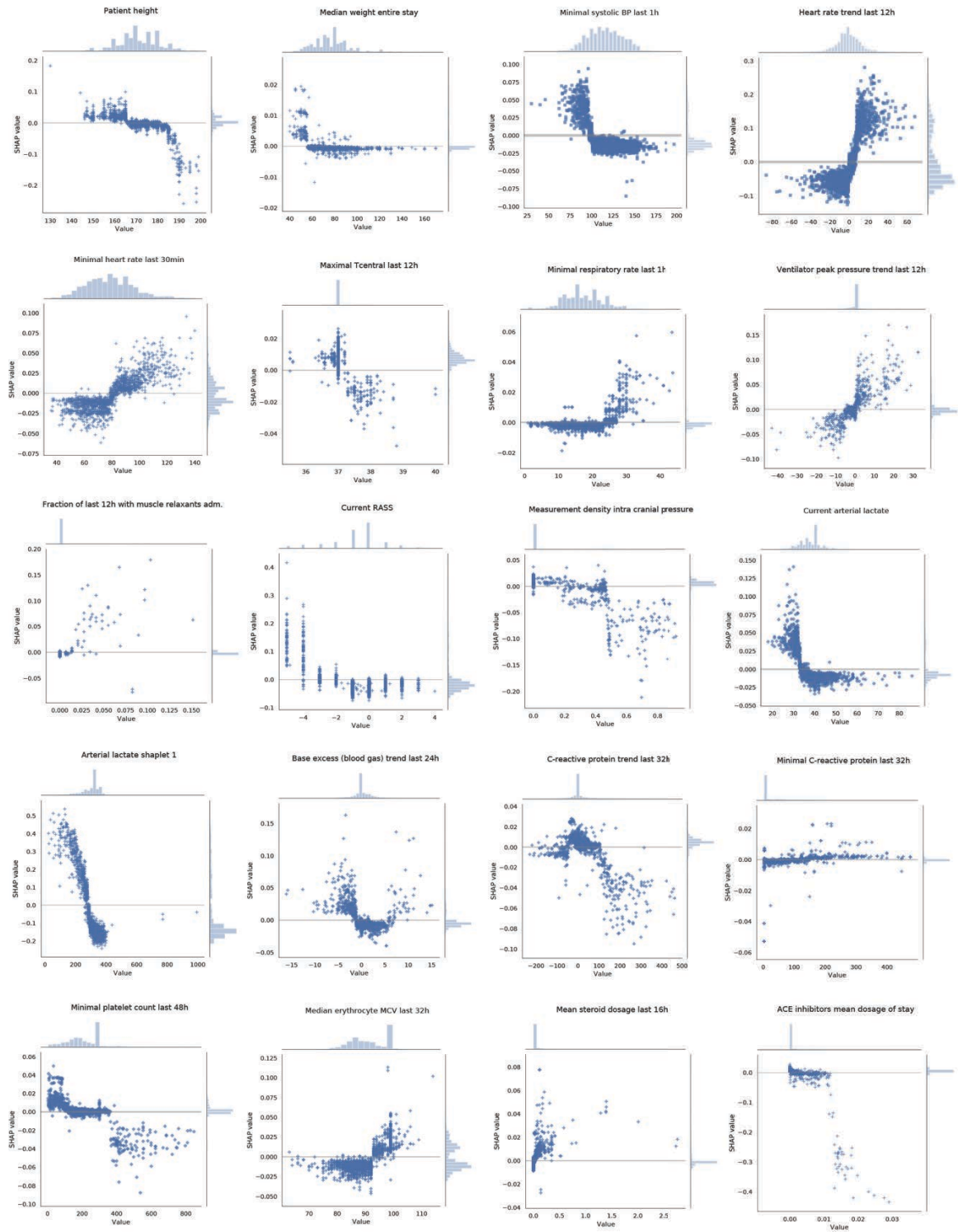

**Supplementary Figure 8: Further examples of *SHAP*-feature relationships.** We show more examples of the relationship between *SHAP* value and feature value. Features were selected from the top500 according to perceived clinical relevance and interpretability.

**Supplementary Figure 9**

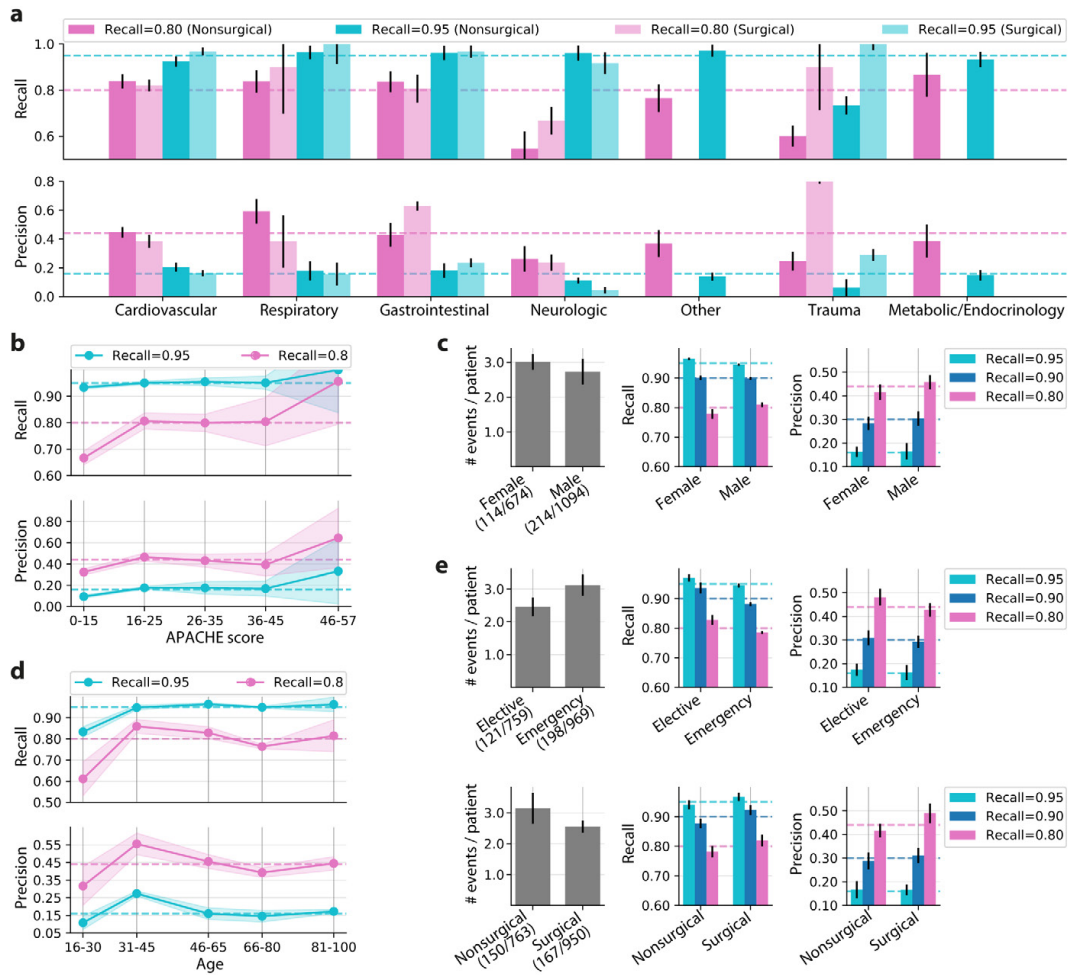

**Supplementary Figure 9: Performance of *circEWS* using different thresholds in different patient sub-cohorts.** In Fig. 4, *circEWS* with 90% recall across the cohort is analysed in terms of its performance in different APACHE patient groups, APACHE severity groups and age groups. This figure analyses the performance of *circEWS* with other recall values, namely 80% and 95%, also with the same patient categorization as Fig. 4 (**a,b,c**). In addition, *circEWS* performance in sub-cohorts categorized by gender and admission types, i.e., elective versus emergency and non-surgical versus surgical, is also analysed (**d** and **e**).

**Supplementary Figure 10**

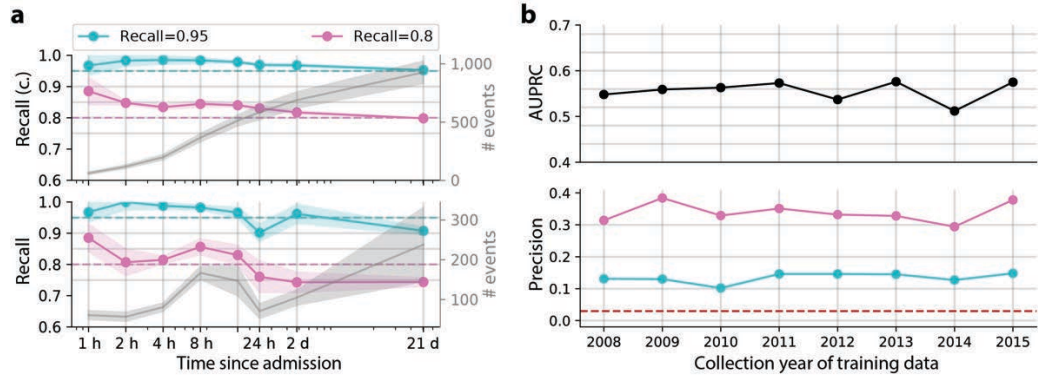

**Supplementary Figure 10: Performance over time of circEWS with different recall rates.** Here we provide analogous analyses to that of Fig. 5, but using circEWS with recall rates of 80% and 95% instead of 90%.

**Supplementary Figure 11**

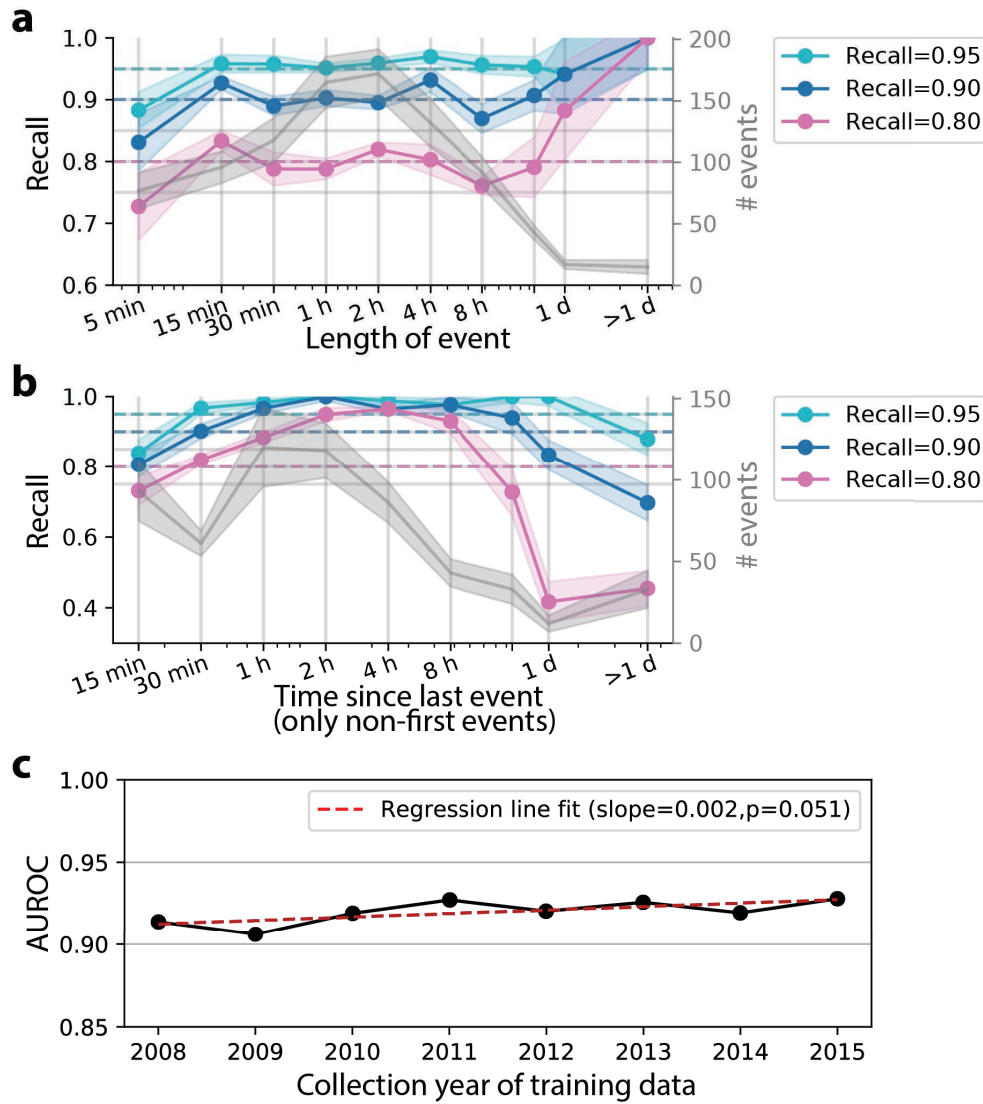

**Supplementary Figure 11. Further analyses. a**: **Performance by event duration.** We report how recall varies for events of differing durations. These events may constitute qualitatively different types of deteriorations, and this variance might manifest in model performance. Unsurprisingly, very short (=5 minutes duration) events had the lowest recall, indicating that these may be spuriously labelled events. The model appears to excel at identifying very long events, however the sample size is low (15 events with duration longer than a day). **b**: **Performance by time since previous event.** We study how the time elapsed between events may affect model performance. Events immediately after other events might be easier to detect, since the patient is likely already close to a state of circulatory failure. This is not confirmed by the data: for very short (< 30 minute) gaps after recovering from a period of circulatory failure, the model exhibits lower recall. **c**: **Other temporal generalization results.** Additional AUROC result for the temporal generalization experiment where the AUROC obtained for training data from different years is shown in black, and a linear regression line fit is shown as a dotted red line. A non-zero slope is significant at 10 % level with two-sided p-value of 0.051 under the Wald test.

**Supplementary Figure 12**

**a**

| Feature type set | AUPRC | prec (rec=0.8) | prec (rec=0.9) | prec (rec=0.95) |
|---|---|---|---|---|
| all | $0.637 \pm 0.027$ | $0.441 \pm 0.024$ | $0.296 \pm 0.024$ | $0.164 \pm 0.028$ |
| - static | $0.630 \pm 0.027$ | $0.422 \pm 0.031$ | $0.300 \pm 0.025$ | $0.168 \pm 0.027$ |
| - measurement intensity | $0.606 \pm 0.030$ | $0.413 \pm 0.026$ | $0.286 \pm 0.028$ | $0.156 \pm 0.022$ |
| - multiresolution summaries | $0.606 \pm 0.021$ | $0.387 \pm 0.026$ | $0.261 \pm 0.026$ | $0.151 \pm 0.034$ |
| - instability history | $0.634 \pm 0.030$ | $0.422 \pm 0.022$ | $0.288 \pm 0.026$ | $0.178 \pm 0.027$ |
| - time to instability | $0.631 \pm 0.026$ | $0.443 \pm 0.026$ | $0.303 \pm 0.025$ | $0.177 \pm 0.019$ |
| - shapelets | $0.632 \pm 0.035$ | $0.434 \pm 0.032$ | $0.268 \pm 0.032$ | $0.181 \pm 0.028$ |
| - entire stay horizon | $0.635 \pm 0.028$ | $0.429 \pm 0.020$ | $0.291 \pm 0.025$ | $0.180 \pm 0.023$ |
| only current values | $0.541 \pm 0.020$ | $0.372 \pm 0.036$ | $0.255 \pm 0.037$ | $0.128 \pm 0.026$ |
| only shapelets | $0.447 \pm 0.063$ | $0.265 \pm 0.031$ | $0.201 \pm 0.013$ | $0.131 \pm 0.005$ |

**b**

| Summary function set | AUPRC | prec (rec=0.8) | prec (rec=0.9) | prec (rec=0.95) |
|---|---|---|---|---|
| location | $0.457 \pm 0.027$ | $0.245 \pm 0.042$ | $0.160 \pm 0.030$ | $0.105 \pm 0.013$ |
| location + trend | $0.519 \pm 0.016$ | $0.328 \pm 0.030$ | $0.220 \pm 0.029$ | $0.128 \pm 0.039$ |
| all | $0.570 \pm 0.020$ | $0.367 \pm 0.034$ | $0.253 \pm 0.028$ | $0.149 \pm 0.020$ |

**c**

| Feature horizon set | AUPRC | prec (rec=0.8) | prec (rec=0.9) | prec (rec=0.95) |
|---|---|---|---|---|
| short term horizon | $0.555 \pm 0.016$ | $0.322 \pm 0.032$ | $0.223 \pm 0.025$ | $0.133 \pm 0.041$ |
| + med term horizon | $0.565 \pm 0.018$ | $0.354 \pm 0.029$ | $0.226 \pm 0.028$ | $0.139 \pm 0.025$ |
| + longer term horizon | $0.568 \pm 0.017$ | $0.362 \pm 0.035$ | $0.256 \pm 0.031$ | $0.148 \pm 0.024$ |
| all horizons | $0.571 \pm 0.019$ | $0.363 \pm 0.034$ | $0.244 \pm 0.024$ | $0.140 \pm 0.021$ |
| only longest term horizon | $0.536 \pm 0.015$ | $0.340 \pm 0.037$ | $0.225 \pm 0.024$ | $0.127 \pm 0.040$ |

**d**

| AUPRC | random | top | min-max |
|---|---|---|---|
| count | $0.36 \pm 0.02$ | $0.37 \pm 0.02$ | $0.33 \pm 0.02$ |
| min | $0.40 \pm 0.05$ | $0.39 \pm 0.02$ | $0.39 \pm 0.03$ |
| distance | $0.41 \pm 0.05$ | $0.41 \pm 0.06$ | $0.41 \pm 0.03$ |
| dist-set | $0.41 \pm 0.02$ | $0.43 \pm 0.04$ | $0.43 \pm 0.02$ |

**e**

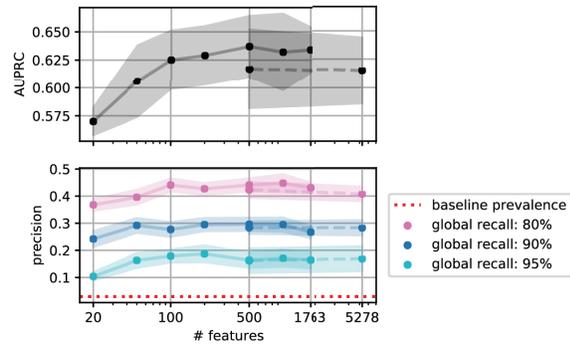

**Supplementary Figure 12: Feature choices. a: Feature ablation by class.** Comparison of model performance when the reference model is ablated by removing individual feature categories. In addition, two baseline models were used. One uses only the 'current' (estimated) values for all variables and includes no special features summarizing the patient history. 'shapelets' used the shapelet mining technique to construct features predictive of circulatory failure. **b: Effect of 5-point summary.** Effect of 5-point summary just using 'location' information, vs. adding trend information (location+trend), vs 'all' possible summaries, including location/variability and trends. **c: Effect of multiple horizons.** Validating the construction principle of the multi-scale history, we check whether the addition of information for medium-term horizon, and the two longest horizons improves the prediction performance. All other features were deactivated so we can study the multi-scale history in isolation. As two baselines we have considered two schemes that do not compute multi-scale information, by just using the shortest, or the longest horizon, on its own, respectively. **d: Shapelet feature types.** All possible shapelet feature extraction methods were compared by training the GBM classifier on the 5 development splits. Selection of statistically significant shapelets had little influence on the results - we therefore picked the min-max procedure because it theoretically guarantees diversity among selected shapelets. More importantly, computing features at multiple timelags (dist-set) outperformed other feature generation approaches. Therefore, we chose to extract min-max dist-set shapelet features for *circEWS*. **e: Effect of number of features.** We studied the effect of the number of features on model performance. We first ranked each feature by its mean absolute *SHAP* value in a model trained on 50% of the data using all 5,278 features. The dashed line indicates results from the model trained on 50% of the data; using the full training data is computationally prohibitive. We marked 1,763 as the number of features whose mean absolute *SHAP* value is non-zero. We saw that more features increase model performance until a saturation point occurs. These results are on the held-out test set, but the decision to use the top 500 features was made on the development test set, which showed qualitatively similar results.

**Supplementary Figure 13**

**a**

| Frequency | Median sampling interval | Short horizon | Medium horizon | Long horizon | Very long horizon |
|---|---|---|---|---|---|
| High | < 15 mins | 30 mins | 60 mins | 240 mins | 720 mins |
| Medium | > 15 mins, < 8 hours | 12 hours | 24 hours | 36 hours | 48 hours |
| Low | ≥ 8 hours | 16 hours | 32 hours | 48 hours | 72 hours |

**b**

| Feature type | Conditions | Horizon lengths |
|---|---|---|
| Fraction of time in sub-event | [MAP ≤ 65 mmHg], [vLac ≥ 2 or aLac ≥ 2 mmol/l], [Dop> 0 μg/min], [Mil > 0 μg/min], [Lev > 0 μg/min], [Theo> 0 μg/min], <br><br>Event L1 := [(vLac ≥ 2 or aLac ≥ 2 mmol/l) and (MAP ≤ 65 mmHg or Dop> 0 μg/min or Mil> 0 μg/min or Theo> 0 μg/min or Lev> 0 μg/min)], <br><br>[Norepb> 0 μg/min and Norepb< 0.1 μg/min], [Norepb≥ 0.1 μg/min], [Epinepb> 0 μg/min and Epinepb< 0.1 μg/min], [Epineph ≥ 0.1 μg/min], <br><br>Event L2 := [(vLac ≥ 2 or aLac ≥ 2 mmol/l) and (Norepb> 0 μg/min and Norepb< 0.1 μg/min or Epinepb> 0 μg/min and Epinepb< 0.1 μg/min)], <br><br>[Vaso> 0 μg/min], <br><br>Event L3 := [(vLac ≥ 2 or aLac ≥ 2 mmol/l) and (Norepb> 0.1 μg/min or Epinepb> 0.1 μg/min or Vaso> 0 μg/min)] | 12 hours, 24 hours, 36 hours, 48 hours, Entire stay |
| Time to last occurrence of sub-event (if any); otherwise set to a large value equal to 30 days if there was no occurence of event. | Same as first row | N/A |
| Current sub-event state | Same as first row | N/A |

**c**

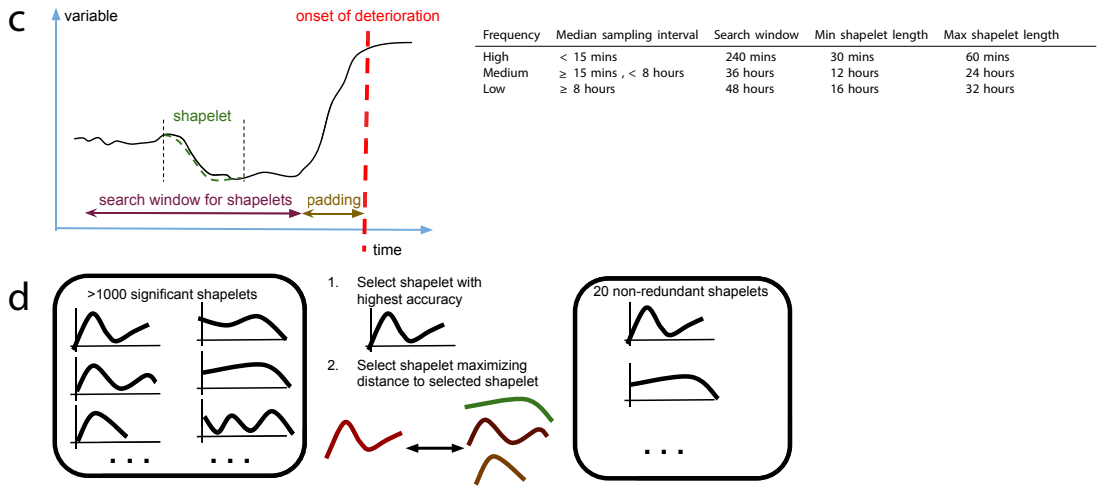

| Frequency | Median sampling interval | Search window | Min shapelet length | Max shapelet length |
|---|---|---|---|---|
| High | < 15 mins | 240 mins | 30 mins | 60 mins |
| Medium | ≥ 15 mins , < 8 hours | 36 hours | 12 hours | 24 hours |
| Low | ≥ 8 hours | 48 hours | 16 hours | 32 hours |

**d**

>1000 significant shapelets

1. Select shapelet with highest accuracy
2. Select shapelet maximizing distance to selected shapelet

20 non-redundant shapelets

**Supplementary Figure 13: Feature details. a: Multiscale feature parameters.** Overview of multi-scale horizon lengths used for the different clinical variables during feature construction, conditional on the length of a typical sampling interval of a variable, shown in the second column. Variables were binned into frequency categories high, medium, and low. For non-pharmaceutical variables the sampling interval was computed from the intervals between observations, and for drugs it corresponded to the effective acting periods compiled using clinical knowledge. **b: List of instability features.** Specification of the instability features computed from specific variables relevant for circulatory failure. Three subtypes of features were constructed, corresponding to rows. The exact specification of individual conditions is listed in brackets in the second column. The last column, in analogy to the multi-scale summaries, displays the horizon lengths over which time-fraction features are computed. Non-adaptive constant horizons were used. For each condition in the first row, time-to-last occurrence and current event state features

were defined similarly. Norepinephrine is abbreviated as noreph. **c: Shapelet extraction details.** The shapelet extraction was performed on time series subsequences that precede onset of circulatory failure where the search window and the size of the shapelet depends on the variable sampling frequency (see Table on right). **d: Shapelet selection details**. The large number of significant shapelets is reduced to a feasible set of 20 shapelets with the min-max selection approach: First, the shapelet with the highest accuracy was selected. Afterwards, shapelets were iteratively added such that the minimal distance to the set of already selected shapelets is maximized.

**Supplementary Table 1**

| Sex | | | |
|---|---|---|---|
| Male | 63.49% | 22919 admissions | |
| Female | 36.51% | 13178 admissions | |
| **Age (years)** | | | |
| Median (Mean) | 66 (62.98) | | |
| Range | 16-98 | | |
| **Length of stay (days)** | | | |
| Median (Mean) | 0.93 (2.09) | | |
| Range | 0-84 | | |
| **Admission type** | | | |
| Emergency | 55.49% | 20032 admissions | |
| Not emergency | 42.49% | 15337 admissions | |
| Unknown | 2.02% | 729 admissions | |
| **Surgical status** | | | |
| Yes | 55.25% | 19944 admissions | |
| No | 42.00% | 15070 admissions | |
| Unknown | 3.00% | 1084 admissions | |
| **APACHE diagnostic group** | | | |
| *Cardiovascular* | | | |
| | Surgical | 23.68% | 8547 admissions |
| | Nonsurgical | 12.96% | 4677 admissions |
| *Neurological* | | | |
| | Surgical | 11.32% | 4088 admissions |
| | Nonsurgical | 17.36% | 6266 admissions |
| *Respiratory* | | | |
| | Surgical | 1.87% | 675 admissions |
| | Nonsurgical | 7.46% | 2694 admissions |
| *Trauma* | | | |
| | Surgical | 0.71% | 258 admissions |
| | Nonsurgical | 4.45% | 1608 admissions |
| *Other* | | 20.17% | 7280 admissions |
| *Unknown* | | 0.01% | 5 admissions |
| **Circulatory dysfunction** | | | |
| Patients with events | 30.60% | 11046 admissions | |
| Mean events per patient with event | 4.16 | | |
| Mean event duration | 320 minutes | | |
| Mean time to first event | 492 minutes | | |
| **Mortality** | 6.11% | 2205 admissions | |
| **APACHE score** | | | |
| Mean (std) | 17.46 (7.86) | | |
| Median (25%, 75%) | 16 (12, 22) | | |
| Range | [0, 57] | | |

**Supplementary table 1: Patient characteristics.** Although stays up to 84 days were observed, we excluded data after 28 days from model development and evaluation.

**Supplementary Table 2**

**a**

| Parameter | Reference model | Reduced model | Baseline |
|---|---|---|---|
| Learning rate | 0.05 | 0.05 | 0.05 |
| Unbalanced mode | TRUE | TRUE | TRUE |
| Frequency of training set sub-sampling per tree | 100 % | 100 % | Never |
| Maximum depth | $\log_2$(Max number of leaves) | $\log_2$(Max number of leaves) | $\log_2$(Max number of leaves) |
| Size of subsample for histogram binning | 1,000,000 | 1,000,000 | 1,000,000 |
| Minimum # samples at leaf | 1000 | 1000 | 1000 |
| Max. # categories for cat. vars to use 1-hot encoding | 100 | 100 | 100 |
| Cat. variable (smoothing, L2) | (0.0,0.0) | (0.0,0.0) | (0.0,0.0) |

**b**

| Parameter | Grid | Reference model optimal | Reduced model optimal | Baseline |
|---|---|---|---|---|
| Maximum number of leaves | 8,16,32,64,128 | 128 | 64 | 128 |
| Fraction of features randomly sampled per tree | 33 %, 66 % | 66 % | 33 % | NA (fixed to 100 %) |
| Fraction of training examples randomly sampled per tree | 33 %, 66 % | 66 % | 66 % | NA (fixed to 100 %) |
| Maximum number of trees (early stopped on validation set) | 5000 | 142 trees (early stopped) | 203 trees (early stopped) | NA (fixed to 1) |

**c**

| Parameter | Grid | Reference model optimal |
|---|---|---|
| Regularization parameter | 0.0001,0.0001,0.001,0.01,0.1,1.0 | 0.01 |
| Loss function | NA | Logistic |
| Penalty | NA | L2 |
| Maximum number of iterations | NA | 1000 |
| Early stopping tolerance | NA | $10^{-3}$ |
| Learning rate schedule | NA | Optimal |
| Class weight in loss function | NA | Balanced |

**d**

| Parameter | | Range | Reference model optimal |
|---|---|---|---|
| Hidden state dimension | temporal | $\mathcal{U}\{1000, 2000\}$ | 1756 |
| | static | $\mathcal{U}\{2, 4\}$ | 3 |
| | joint | $\{2^\alpha : \alpha \in \mathcal{U}\{4, 7\}\}$ | 16 |
| Learning rate | | $\{10^\gamma : \gamma \in \mathcal{U}(-5, -2)\}$ | $1.2 \times 10^{-4}$ |
| Dropout | | $\mathcal{U}(0.2, 0.8)$ | 0.38 |
| Activation function | | {relu, tanh, sigmoid} | tanh |

**Supplementary Table 2: Hyperparameters. a: Fixed hyperparameters for tree-based models.** Fixed hyperparameters used for tree-based (ensemble) models implemented using the *lightGBM* library. Full and compact models used the same hyperparameters, as well as the decision tree baseline, except that no random sub-sampling of the training set was used in the latter case. **b: Hyperparameter grid and optimal values for tree-based models.** Hyperparameter search space for the grid search for tree-based models implemented using the *lightGBM* library, yielding a grid size of 20. Decision trees had a reduced grid because feature/training set sub-sampling is not applicable for them. The last row shows the effective number of trees that are added to the ensemble, using early-stopping on the validation set, for the full/compact model respectively. **c: Fixed/grid hyperparameters and optimal values for LR.** Hyperparameter search grid and fixed hyperparameters for the LogReg model. Only the regularization parameter weighting the L2 weight decay was included in the grid. The 'Optimal' learning rate schedule is based on a heuristic by Leon Bottou, and the early stopping tolerance refers to the minimal improvement on the training/validation loss that is achieved in every epoch until the optimization is early stopped.

**d: Hyperparameter range and optimal values for LSTM.** Random search was used to search for the optimal set of hyperparameters with the specified range for the LSTM models, and 16 sets of hyperparameters were randomly drawn. Early stopping based on the validation results was also used during the model training.

## Supplementary Table 3

| Timestamp artifact description (of a variable for a patient) | Solution |
|---|---|
| The *SampleTime* and the *EnterTime* differ in the "year" but not in the "month" field. | Replace the value of the "year" field in the *SampleTime* with that of the *EnterTime*. |
| The *SampleTime* and the *EnterTime* differ in both the "year" and the "month" fields, and the "year" of the *EnterTime* is the same as the "year" of the *SampleTime* in the next record. | Delete all records with *SampleTime* earlier than *AdmissionTime*. |
| The *SampleTime* and the *EnterTime* differ in both the "year" and the "month" fields, and the *SampleTime* of the current and the next record only differs in the "month" field, but the *SampleTime* of the current record and the *EnterTime* of the previous/next record has the same "year". | Replace the value of the "year" field in the *SampleTime* with that of the previous/next record. |
| The *SampleTime* and the *EnterTime* differ in the "month" not in the "year" field, and the absolute difference in "day" is smaller than 2 days. | Replace the value of the "month" field in the *SampleTime* with that of the *EnterTime*. |
| The *SampleTime* and the *EnterTime* differ in the "month" not in the "year" field, and the absolute difference in "day" is larger than 1 day, but the *SampleTime* of the current and the previous/next record has the same "day". | Replace the value of the "day" field in the *SampleTime* with that of the previous/next record. |
| The absolute difference in "month" between the *SampleTime* and the *EnterTime* is 11, and the *SampleTime* and the *EnterTime* is the same in "month" in the previous/next record. | Add/subtract one from the "year" value of the *SampleTime* of the record. |
| The "month" and "day" of *SampleTime* are swapped compared to *EnterTime*. | Swap "month" and "day" of *SampleTime* according to *EnterTime*. |
| Uncorrectable *SampleTime* resulting in a long time gap at the end of the stay, and the number of records after the gap is less than 5 % of the total number of records. | Delete all records after the gap. |
| The units digit of the "month" of the *SampleTime* is swapped with the tens digit of the "day" of the *SampleTime*. | Swap back the units digit of the "month" of the *SampleTime* with the tens digit of the "day" of the *SampleTime*. |
| The span of *SampleTime* of the records before the gap is larger than a day while the span of the *EnterTime* of the records is smaller than 6 hours, i.e. the records are history long before the admission. | Delete all records before the *AdmissionTime*. |
| The *SampleTime* and the *EnterTime* only differ in the tens digit of the "day" field, i.e. the difference in "day" is a multiple of 10. | Replace the value of the "day" field in the *SampleTime* to that of the *EnterTime*. |
| The date of the *EnterTime* is the end of a month X but the date of the *DateTime* is the beginning of the month X-1 or X. | Replace the "month" of the *DateTime* with X+1. |
| The *SampleTime* of the last record before the gap is before the *AdmissionTime* (or the *SampleTime* of the first monvals record for non-monval tables), and the total number of records before the gaps is less than 10 % of the total number of records. | Delete all records before the gap. |
| The *SampleTime* of the first record after the gap is after the *SampleTime* of the last monvals record for non-monval tables, and the total number of records after the gaps is less than 10 % of the total number of records. | Delete all records after the gap. |

**Supplementary Table 3:** Description of time-point artefacts and corresponding solutions in HiRID.

**Supplementary Table 4**

| Forward variable selection rank (std) | Variable | Marginal AUPRC improvement |
|---|---|---|
| 1 (0.0) | Lactate | +0.29 |
| 2 (1.2) | MAP | +0.109 |
| 3 (7.0) | Non-opioid analgesics | +0.028 |
| 4 (4.3) | Peak inspiratory pressure (Ventilator) | +0.006 |
| 5 (5.2) | Supplemental oxygen | +0.009 |
| 6 (3.5) | Time since ICU admission | +0.001 |
| 7-10 (3.6, 1.9, 1.9, 3.2) | Dobutamine, Milrinone, Levosimendan, Theophylline | None |
| 11 (4.8) | C-reactive protein | None |
| 12 (5.6) | RASS | None |
| 13 (3.0) | Cardiac output | +0.003 |
| 14 (2.4) | Serum glucose | None |
| 15 (7.1) | Diastolic BP | None |
| 16 (3.8) | SpO$_2$ | None |
| 17 (5.6) | Heart rate | None |
| 18* (3.2) | Patient age | +0.004 |
| 19 (5.0) | INR | None |
| 20 (6.0) | Systolic BP | None |

.

**Supplementary Table 4: Forward variable selection.** Shown is the rank (column 1) of a variable (column 2) and performance increase (column 3) in forward variable selection. Point estimates of the ranks were obtained on the held-out split and the standard deviations of the ranks were obtained by running the same procedure on the five development splits. The performance was measured using the AUPRC (time-slice based evaluation) on the validation set. A lightGBM model with fixed hyperparameters was used (at most 64 leaves per tree, and column/row sampling ratios of 75% per tree, using early stopping after 20 epochs of no change). The third column indicates the AUPRC improvement that was obtained by adding this variable to the best model obtained so far, only using variables in rows above the variable. 'None' indicates that no improvement resulted from the addition of the variable. The best model obtained by this procedure contains 18 variables (and the last variable included in the model is marked with *).

**Supplementary Table 5:**

List of variables included (and # of features per variable)
in full model (top 500 features).

Supplementary excel sheet "circEWS_Variables_Overview_Table.xlsx"

**Columns in this spreadsheet**
- Variable name
- #features derived from this variable
- Inclusion of variable in compact model
- Relevant MIMIC identifiers
- Default value
- Permitted range of values
- Type (continuous, binary, categorical)

**Further sheets**
- List of drugs comprising combined drugs
- List of variables formed from merging other variables (non-drugs)

**Supplementary Table 5: List of variables in *circEWS* model.** This table lists the full set of variables used in our models. The "full" model uses 112 variables associated with the top 500 features. The "compact" model uses only features from the top-20 variables, excluding those not identifiable in MIMIC, resulting in 17 variables generating 176 features. Where relevant, we further provided the chosen mapping to MIMIC ITEMIDs, as well as the default values used in imputation (selected using prior clinical knowledge), and the range of permitted values for artefact removal.